\pdfoutput=1

\documentclass[11pt]{article}

\usepackage{acl}

\usepackage{times}
\usepackage{latexsym}

\usepackage[T1]{fontenc}

\usepackage[utf8]{inputenc}

\usepackage{microtype}

\usepackage{inconsolata}

\usepackage{listings}
\usepackage{graphicx}
\usepackage{listings}
\usepackage{pythonhighlight}
\usepackage{soul}
\usepackage[many]{tcolorbox}    			    
\usepackage{setspace}               
\usepackage{multicol} 
\usepackage{booktabs}
\usepackage{enumitem}
\usepackage{outlines}
\newtcolorbox{boxA}{
    boxrule = 1.0pt,
    colframe = black, 
    lefthand width=0.2em,
    colback = white,
    left=0em,
    top=0em
}

\title{ConSiDERS-The-Human Evaluation Framework:\\  
Rethinking Human Evaluation for Generative Large Language Models}

\author{Aparna Elangovan,
  Ling Liu,
  Lei Xu,
  Sravan Bodapati,
  Dan Roth \\
  AWS AI Labs \\
  \texttt{\{aeg,lingliun,leixx,sravanb,drot\}@amazon.com}
  }

\begin{document}

\maketitle
\begin{abstract}
In this position paper, we argue that human evaluation of generative large language models (LLMs) should be a multidisciplinary undertaking that draws upon insights from disciplines such as user experience research and human behavioral psychology to ensure that the experimental design and results are reliable.  The conclusions from these evaluations, thus, must consider factors such as usability, aesthetics, and cognitive biases. We highlight how cognitive biases can conflate fluent information and truthfulness, and how cognitive uncertainty affects the reliability of rating scores such as Likert. Furthermore,  the evaluation should differentiate the capabilities and weaknesses of increasingly powerful large language models -- which requires effective test sets. The scalability of human evaluation is also crucial to wider adoption. Hence, to design an effective human evaluation system in the age of generative NLP, we propose the \textbf{ConSiDERS-The-Human} evaluation framework consisting of 6 pillars -- \textbf{\ul{Con}}sistency, \textbf{\ul{S}}coring Cr\textbf{\ul{i}}tera, \textbf{\ul{D}}ifferentiating, User \textbf{\ul{E}}xperience, \textbf{\ul{R}}esponsible, and \textbf{\ul{S}}calability.

\end{abstract}

\section{Introduction}
Generative tasks in natural language processing (NLP) have to rely on human evaluation, as the current set of automated metrics does not correlate well with human judgment \cite{gao-wan-2022-dialsummeval, deutsch-etal-2022-examining}. Human evaluation tends to be expensive and difficult to repeat or reproduce \cite{belz-etal-2023-non, belz-etal-2020-disentangling}. Even more importantly, an all-too-common scenario tends to be that the evaluation method is fundamentally misaligned with the problem statement \cite {hamalainen-alnajjar-2021-great}. In the age of generative large language models (LLMs) with increasing capabilities that can generate fluent content to even fool humans \cite{clark-etal-2021-thats}, ensuring that the human evaluation is set up appropriately to measure the right aspects and reach the right conclusions is crucial.

In this position paper, first, we argue that to design and interpret the results of human evaluation accurately, the evaluation pipeline needs to be \textbf{human-centric} in the age of generative AI, accounting for human evaluators and their cognitive biases. The field of user experience (UX) takes into account the emotional states of a user, a.k.a. \textit{how a user feels} \cite{Marques2021-ei,hartson2012ux}. It is a well-known fact in UX that users tend to be heavily influenced by aesthetic aspects, while actual function or usability aspects take a second place when users perceive a system as useful, leading to the notion ``\textit{what is beautiful is useful}'' \cite{SONDEREGGER2010403, TUCH20121596, Hamborg2014}. Aesthetics also extends to language. Factors such as fluency can affect the evaluation, outweighing the actual content or substance \cite{10.1093/acprof:oso/9780199732142.003.0055}. Studies in human-computer interface (HCI), cognitive, and social psychology have demonstrated that processing fluency -- the ease with which information is perceived and processed in the human mind -- has a positive effect on evaluation \cite{10.1145/3544549.3585739, tsai2011does, greifeneder2017interplay}. Current state-of-the-art (SOTA) LLMs tend to be quite fluent and produce content that is easy to read and understand, and as a result, users can conflate fluency and usefulness. Therefore, we need to closely examine our human evaluation before reaching conclusions such as -- \textit{The LLM can perform function <x> similar to or better than a trained professional}. NLP evaluation procedures, therefore, at the very least must delineate style vs.\ substance.

Secondly, the \textbf{effectiveness of the test set} in measuring the capabilities of a model is critical, as ineffective test sets cannot adequately evaluate these models, a common theme that has surfaced in many leader-boards and public data sets \cite{tedeschi-etal-2023-whats,elangovan-etal-2021-memorization}. 

Hence, in this position paper, we make the following contributions: 

\textbf{1)} We propose a framework, a structure for organizing and contextualizing human evaluation, that can be customized and adapted to specific contexts. Our proposed framework --  the \textbf{ConSiDERS-The-Human} evaluation framework -- has 6 pillars:

\begin{boxA}{\textbf{The 6 pillars of ConSiDERS-The-Human Evaluation Framework:} (See \textit{Checklist} in Appendix~\ref{sec:checklist} to follow.)}
\begin{itemize}[topsep=1pt]
\setlength{\itemindent}{-1em}
\setlength{\leftmargin}{0.0cm}
\addtolength\itemsep{-3mm}
    \item \textbf{Con}sistency of human evaluation: The findings of human evaluation must be reliable and generalizable.
    \item \textbf{S}coring Cr\textbf{i}teria: The scoring criteria must include both general purpose criteria such as readability, as well as be tailored to fit the goal of the target tasks or domains.   
    \item \textbf{D}ifferentiating: The evaluation test sets must be able to differentiate the various capabilities as well as the weaknesses of generative LLMs. 
    \item User \textbf{E}xperience:  The evaluation must take into account user experience, including their emotions \& cognitive biases, when designing experiments and interpreting results.
    \item \textbf{R}esponsible:   The evaluation needs to account for responsible AI  including aspects such as bias, safety, robustness, and privacy capabilities of the model. 
    \item \textbf{S}calability:  Human evaluation must be scalable for pragmatic widespread adoption. 
    
\end{itemize}
\end{boxA}



\textbf{2)} We make the case for why UX and the psychology of cognitive biases should be at the forefront of human evaluation. In the last 20 years, less than 7\% of the papers (only 16 papers) with ``human'' and ``eval'' in their title available in ACL Anthology mention user experience-related keywords in either the title or the abstract (see query in Appendix~\ref{app:sec:humanevalkeywordintitlewithusability}). 

\textbf{3)} We highlight how neglecting the role of cognitive biases in human evaluation can lead to incorrect inclusions from the study. We, therefore, provide specific recommendations to mitigate the effects of common cognitive biases. We also provide tips to troubleshoot and improve consistency issues in human evaluation.

In the rest of this paper, we introduce the necessary background concepts in Section~\ref{sec:background} and explore each of the 6 pillars in detail in Section~\ref{sec:framework}.

\section{Background concepts}\label{sec:background}

\subsection{ Usability}\label{sec:conceptuxusablity}
Usability, according to ISO 9241--11:2018, is defined as -- ``\textit{The extent to which a system, product or service can be used by specified users to achieve specified goals with effectiveness, efficiency, and satisfaction in a specified context of use}'' \cite{barnum2020usability}. Usability testing includes the following five elements or the 5 Es of usability \cite{niranjanamurthy2014research, barnum2020usability}. \textbf{1) Easy to learn:} This aims to address questions such as a) How easy is it for users to complete basic tasks the first time they use the system? b) When users return to the design after a period of not using it, how well is the user able to recollect how to use the system? \textbf{2) Efficiency}: How quickly can experienced users accomplish tasks? \textbf{3) Effective}:  How completely and accurately the work or experience is completed or goals reached? \textbf{4) Error tolerant}: How many errors do users make, how critical are these errors, and how easily can they recover from the errors? \textbf{5) Engaging / Satisfaction}: How much does the user like using the system? These 5Es are crucial when designing human evaluation solutions to obtain reliable evaluation results.

\subsection{UX and HCI}\label{sec:conceptux}

User experience (UX), according to ISO 9241-110:2010  is defined as -- ``\textit{a person’s perceptions and responses that result from the use and/or anticipated use of a product, system or service}'' \cite{10.1145/2702613.2732511}. UX, thus, expands beyond the concepts of the 5Es of usability to take into account the broader emotions experienced by the users when using the system  \cite{Marques2021-ei, hartson2012ux}. In other words, UX takes into account usability as well as the users' feelings as to how products ``\textit{dazzle their senses, touch their hearts and stimulate their minds}'' \cite{Marques2021-ei}. HCI is the UX when humans interact with computer systems, including user interfaces and how information is presented on the digital screen. Commercial LLMs facing end users, such as ChatGPT, have dazzled the minds of their users. Users' emotion, including perceived usability or usefulness,  tends to be heavily influenced by aesthetics \cite{hartson2012ux}. Aesthetics influences the user heavily initially or in one-off tests, but over time aesthetics plays a much lesser role and usability becomes crucial \cite{doi:10.1080/00140139.2012.672658}. The field of UX, therefore,  attempts to disambiguate the perceptions that users form as a result of aesthetics, and the need to measure actual function as defined by the 5Es of usability, whilst embracing user emotions.

\subsubsection{Measuring UX: Perception vs. Performance}
UX feedback can be qualitative such as user interviews or quantitative metrics. Quantitative UX metrics include performance-based metrics such as time to complete a task, the errors the users encounter, and \textbf{perception-based self-reported metrics} through rating scale and preferences \cite{TULLIS201341}. Likert scale is a type of attitude scale, a special case of a rating scale, that measures the degree to which a person agrees/disagrees with a given statement \cite{RePEc:hal:journl:hal-02557308}. In NLG, performance-based metrics designed to measure the impact of the end system are considered extrinsic evaluations, while intrinsic evaluations attempt to evaluate properties of the NLG text \cite{van-der-lee-etal-2019-best}. Measuring aspects such as fluency are intrinsic evaluations, usually measured through a rating scale or preference tests, where the evaluator is asked which model's output they prefer. \ul{Rating scales and preference tests are based on user perception}, and therefore subject to cognitive biases. 

\subsection{Cognitive biases}\label{sec:cognitivebias}
Cognitive biases are systematic errors in human judgment or aspects that drive irrational behavior \cite{doi:10.1126/science.185.4157.1124, Ellis2018}. This is usually a result of relying on heuristics to  make a decision. There are many types of cognitive biases, the sources can be broadly categorized into \textbf{a)} too much information \textbf{b)} lack of information or lack of understanding or meaning associated with the information \textbf{c)} need to act or make judgments fast \textbf{d)} information that is remembered or recalled \cite{10.1145/3406522.3446023}. For example, when a user is presented with a long list (too much information) during information  retrieval, to quickly filter out information, the user would simply click on the first link due to position bias \cite{10.1145/3406522.3446023}.  

There are several studies on the effects of cognitive biases on information search and retrieval \cite{10.1197/jamia.M2411, white2013beliefs} and crowdsourcing \cite{10.1145/3159652.3159654, 10.1145/3313831.3376318}. For instance, \citet{white2013beliefs} finds that evaluation of search and retrieval systems is impacted by confirmation bias -- people’s unconscious tendency to prefer information that confirms their beliefs and disregard evidence that refutes it. There are over 180 different types of cognitive biases identified \cite{10.1145/3406522.3446023, doi:10.1126/science.185.4157.1124}, resulting from a range of factors from how questions are framed  \cite{Choi2005-rs} to prior beliefs \cite{white2013beliefs} that attempt to explain the heuristics humans use to make decisions. These heuristics also impact how humans evaluate LLMs.

\section{ConSiDERS-The-Human framework}\label{sec:framework}

\subsection{User \textit{\underline{E}}xperience} \label{sec:ExperimentalDesign}
Cognitive biases play a key role in how humans judge or rate the system. Despite this, there is little reporting of the influence of these biases in NLG tasks with human evaluation \cite{schoch-etal-2020-problem}. In this section, we make the case as to why UX and the psychology of cognitive biases are crucial components of human evaluation in NLP. Since UX and the psychology of cognitive biases are entire fields on their own, it is impossible to cover all the details in this paper. We highlight the significant impact of cognitive biases on human evaluation of NLG tasks that can lead to incorrect conclusions. 

\textit{Processing fluency} -- the ease with which information is processed by the human mind \cite{tsai2011does} -- affects factors such as \underline{perceived} truthfulness and usefulness of statements. These lessons from psychology also apply to NLG human evaluation, where the human evaluation strategy needs to isolate the effects of linguistic fluency vs. aspects such as factual correctness. As presented in the introduction section, the notion ``\textit{what is beautiful is useful}'' \cite{SONDEREGGER2010403, Hamborg2014} also extends to language, where information that is presented in an easy-to-process manner can be perceived as true \cite{10.1162/daed.2006.135.2.136}. To be clear, we are not calling for linguistic fluency and coherence to be trivialized. On the contrary, we highlight how powerful its influence is on human judgment and evaluation. In the following section~\ref{sec:mitigatecognitivebias}, we provide a few tips on how to isolate such effects in NLG evaluation.  




\subsubsection{Recommendations to mitigate common cognitive biases in NLG evaluation}\label{sec:mitigatecognitivebias}

\textbf{1. Cognitive uncertainty in user feedback including rating schemes:} Explicit user feedback such as 1-5 rating scales, and preference-based tests are inherently subject to cognitive uncertainty, therefore the same user can change their rating on the same item when asked again at a later point in time, even within a few minutes after the initial rating \cite{doi:10.1080/0144929X.2019.1604804, Kotkov2022RatingCI, 10.1007/978-3-642-02247-0_24}. Uncertainty in user feedback is a well-known problem in recommendation systems, where a user's rating is considered to be noisy \cite{hill1995recommending}. \citet{doi:10.1080/0144929X.2019.1604804} demonstrate the scale of the problem where 65\% of the users change their rating even within short intervals between re-rated items. Intra-user rating consistency tends to be higher when the ratings are extreme, e.g., very good or very bad, whereas the in-between ratings tend to have lower consistency \cite{10.1007/978-3-642-02247-0_24, 10.1145/1639714.1639744}. Hence, cognitive uncertainty is more likely to affect model evaluations where the outputs are neither very good nor very bad, where the ratings fall in the mid-point, e.g., 3 on a rating scale of 1 to 5. Similar problems can also surface in preference tests, where users choose ``A is marginally better than B''  when they don't have strong opinions.

\textit{Mitigation}: Rating denoising algorithms in recommendation systems obtain multiple ratings from the same user and attempt to keep only some of the ratings \cite{JOORABLOO2022242, 10.1145/1639714.1639744}. For instance, \citet{10.1145/1639714.1639744} propose to keep only those intra-user ratings whose difference is less than a predefined threshold and choose the mildest rating (most neutral rating) as the final rating for that user. The intuition here is that the mildest rating is not likely to affect the items recommended to the user. NLP human evaluation can potentially leverage such denoising algorithms. Some studies have reported that binary preference-based evaluation has \textit{better consistency} compared to rating \cite{belz-kow-2010-comparing}.

\textbf{2. Conflate fluency with attributes such as truthfulness:}  In psychology, the subjective ease with which the mind processes information is more likely to be judged as true \cite{KOCH2012481, 1999-11900-005}. This subjective ease with which the mind processes information can be due to factors including how information is presented, how frequently it is repeated and cues of familiarity, such as native speakers can seem truer than those with a foreign accent \cite{doi:10.1146/annurev-psych-010419-050807}. Thus, humans rely on shortcuts and draw inferences on aspects such as truthfulness from \ul{feelings}. Conflating fluency with truthfulness is a result of a cognitive bias called the halo effect. The halo effect is the influence of a global evaluation on evaluations of individual attributes \cite{halo}. This particularly affects NLP scoring criteria such as factual completeness, salience, and truthfulness, where these individual attributes can be impacted by the overall global attribute -- linguistically fluent easy to process information. Therefore, the reliability of rating schemes particularly affects tasks that require rigorous detailed inspection, such as truthfulness or factual completeness. This cognitive bias has been largely ignored when using rating scales including Likert to evaluate the factual correctness of LLMs. As a result, experimenters can inadvertently conclude that LLMs are as \textbf{factually comprehensive} in performing a given function as a trained professional like a doctor. We find that 9 out of the 19 papers we sampled from top-tier medical journals use the perception-based Likert scale to evaluate factual correctness and completeness (details in Appendix~\ref{app:sec:chatgpt}), indicating how widespread these practices are. 

\textit{Mitigation}: Tasks such as fact-checking and factual completeness that require inspecting individual traits must be ideally split into atomic facts as detailed in  Section~\ref{sec:consistency} to isolate the impact of fluency and ease of information processing vs. facts.

\textbf{3. Over-reliance on initial information:}  Anchoring bias or over-reliance on an initial piece of information \cite{doi:10.1126/science.185.4157.1124, FURNHAM201135} affects how models are scored using preference or rating tests. For instance, when performing preference tests, if the model presented first on the screen is always the same, then the initial perception of evaluators can have a significant impact on the rest of the evaluation. 

\textit{Mitigation}: It is important to shuffle the display order so that the order, such as the sequence of human evaluation tasks, doesn't give away the model. In preference tests, within each task where the outputs of say 2 models (e.g., model A vs. B) are compared, the underlying model representing A and B must also be randomly shuffled so that A does not always refer to the same model.

\textbf{4. Perception  vs. Performance}  
Most self-reported feedback such as user ratings tends to be based on \textbf{human perception} \cite{TULLIS201341}. While perception-based metrics represent how a user feels are necessary to measure subjective aspects such as readability, they do not capture functional performance-based metrics such as efficiency. Studies have shown that users can dislike a system that performs well or like a system that does not perform well \cite{doi:10.1177/154193129303700406/RobertWBailey}. For instance, in an experiment where participants are asked to choose between one-level and two-level menus for sorting categories, participants preferred the two level menu, even though during actual use the one level menu was much faster and less error-prone \cite{doi:10.1177/154193129303700406/RobertWBailey, HAYHOE1990677}. 
This demonstrates the discrepancy between preferences reported vs. measurements of efficiency. 

In the context of LLM evaluation, consider a hypothetical application scenario where an LLM generated output is used to automatically draft email responses. In this scenario, how much a user prefers the output of a model versus the reality of how useful that model output is in boosting productivity (time spent drafting emails) need not be correlated. The main challenge with conducting usability studies is that the software system needs to be built to study how it impacts an end user’s productivity. In addition, confounding factors such as poorly designed UI can result in reducing productivity and these aspects need to be taken into account when designing and analyzing usability studies to understand the impact of an LLM generated output. \citet{liebling-etal-2022-opportunities} highlight a similar problem in evaluating large-scale machine translation, where the end user's experience can be quite different from simply evaluating the model output.

\textit{Mitigation:} NLP tasks that are meant to assist end users, must eventually \ul{conduct usability studies}  to capture performance metrics such as efficiency.



\subsection {\underline{\textit{Con}}sistency of human evaluation}\label{sec:consistency}
Reproducibility of experiments in science is a widespread challenge and has even led to the term ``replication crisis'' being coined \cite{Baker2016}. Human evaluation is no exception to this challenge, where less than 5\% of human evaluations are repeatable \cite{belz-etal-2023-non}. Despite the challenges, consistency cannot be ignored, as poor reproducibility can point to core design problems. 


Broadly speaking, \textbf{non-random or systematic inconsistencies} primarily arise due to 5 main design flaws \textbf{1)} ill-defined evaluation guidelines provided to the annotators \textbf{2)} high complexity task \textbf{3)} evaluators who are not well qualified or suited to the task \textbf{4)} small number of evaluators and/or evaluation set size \textbf{5)} rating scales such as the Likert. We specifically need to be able to differentiate between random and non-random inconsistencies, as human evaluators are subject to decision errors/outcomes depending on their cognitive state. Random errors can neither be predicted nor controlled \cite{8634261_Unbiased_Replication_Studies}. Thus, understanding the role of non-random variations due to system design is key to improving the consistency of evaluation.

\textbf{1) Ill-defined or complex evaluation guidelines:} Ill-defined guidelines are often ambiguous, incomplete, do not address boundary cases, and do not provide adequate examples \cite{Gadiraju_10.1145/3078714.3078715}. To illustrate this point, \citet{Pradhan2022-wq} use the example of a seemingly simple task ``\textit{Is there a dog in this image?}'' where the authors point out how even this simple task can elicit several clarification questions such as ``Does the dog need to be a real animal?'', ``What if the dog is only partially visible in the image?'' and ``What about a wolf?''. The authors further suggest a 3-stage workflow to improve annotation guidelines: Stage 1 involves workers identifying ambiguous samples; Stage 2 involves labeling a few ambiguous examples to add as clarifying examples in the instructions; and Stage 3 involves workers performing the actual annotation using the revised guidelines with the clarifying examples. Overly long annotation guidelines might even require training the annotators, and hence annotators with task-relevant experience tend to be referred \cite{rottger-etal-2022-two}. \citet{Wu_Quinn_2017} find that using simple vocabulary and logical ordering of instructions can improve the guideline quality. Improving guidelines alone may not be sufficient, as enhancing the user interface design can help reduce cognitive load of the annotator, which in-turn can improve the accuracy of annotation tasks \cite{Sampath10.1145/2556288.2557155}.

A simple way to identify deficiencies in guidelines is to have ``experts'' independently evaluate a set of task items using the guideline and compute the IRA score using an appropriate metric such as a Kappa score. Low IRA can indicate potential problems with the annotation guidelines, in which case revising the guidelines iteratively can lead to reasonable agreement \cite{iskender-etal-2021-reliability}.  

The 5Es of usability testing criteria listed in Section~\ref{sec:conceptuxusablity} is an important strategy to follow when designing human evaluation solutions, including aspects such as how quickly human evaluators can complete their tasks while minimizing errors, how easily the annotation guidelines can be followed and memorized. For example, an overly detailed hard-to-remember evaluation guideline can simply result in poor usability affecting ease of learning, efficiency, and error tolerance resulting in poor inter-rater agreement and/or very slow evaluation turnaround times.

\textbf{2) High task complexity:} Tasks that involve high cognitive load for the evaluator, can lead to lower agreement \cite{kim2023interannotator, Pommeranz2012, liu-etal-2023-revisiting}. High cognitive load can even simply involve asking the evaluator to assign a rating of 1-5 \cite{freitag-etal-2021-experts}. One way of mitigating this is to simplify the task. Simplification is key to obtaining consistent results. 

An example of task simplification to evaluate the factual completeness or saliency of a model-generated summary is to break a long text into hierarchical units of facts \cite{liu-etal-2023-revisiting} using protocols such as Pyramid \cite{nenkova-passonneau-2004-evaluating}. Breaking a reference summary into atomic facts allows the evaluators to verify if the fact is present, and the total number of facts captured in the summary can be summed up automatically to compute the overall score. Thus, the fact-level recall score is likely to yield much more consistent and reliable results compared to a grading scheme, such as the Likert scale, which would involve asking the evaluator to rate the completeness, framing the problem as ``\textit{How complete do you think the summary is on a scale of 1-5}''.  A flip side to breaking long text into atomic units is that it can lead to loss of information when measuring certain types of criteria, e.g., qualitative aspects such as  coherence cannot be evaluated using atomic facts.

\textbf{3) Ill-suited evaluators:} Ill-trained annotators can also be a source of inconsistency, usually a scenario encountered when using crowdsourced workers to evaluate specialized tasks. Annotators who demonstrate poor attention can be identified using a set of attention check questions \cite{Agley2022}. Similarly, ill-trained or ill-qualified workers can be identified using an ``exam set'', a test set for which answers are known. Lower IRA can also be a result of variation in the skills or qualifications of the evaluators \cite{Artstein2017}. 

\textbf{4) Small number of evaluators and/or test set:} Low number of participants, or the sample size of the evaluators, is one of the key contributors to poor reproducibility \cite{Maxwell2015-bi,  Button2013}. Using a larger pool of evaluators can mitigate experimenter bias introduced when using very small groups \cite{8634261_Unbiased_Replication_Studies}. A small sample size of the test set is another source that introduces replication problems,
A caveat here is that a large group of evaluators and a large test set is necessary but not sufficient to ensure that the study is experiment bias-free, as selection bias can result in a large size that is not representative of the target population \cite{Kaplan2014-dq}.

\textbf{5) Rating scales such as the Likert:}   As discussed in Section~\ref{sec:mitigatecognitivebias}, self-reported user feedback using rating schemes is inherently noisy and, therefore, unreliable. Despite 50\% of human evaluations relying on  Likert scale \cite{VANDERLEE2021101151}, there has been little investigation into the controversies surrounding it in NLP. These include Likert's consistency issues \cite{doi:10.1080/01488376.2011.580697}, aggregation \& interpretation of scores \cite{Bishop2015-dk, willits2016another} and the methods to compute IRA \cite{10.3389/fpsyg.2017.00777} that the NLP community needs to research further.


\subsubsection {IRA: Importance and caveats}

IRA or inter-rater agreement in human evaluation measures how well two or more evaluators agree on the scores or preferences they assign independently. While it is important to measure IRA to detect problems in the design, especially given that only 18\% of the papers using human evaluation report IRA \cite{amidei-etal-2019-agreement}, there are several caveats called out on the use of IRA metrics in the medical community which we will be discussing below. 

\citet{GISEV2013330} guide when to use which IRA measure, e.g., Krippendorff's-$\alpha$ vs.\  Cohen's-$\kappa$, depending on experiment design factors such as the number of annotators and the type of variable (e.g., ordinal, nominal, etc). Despite such high-level guidance,  which IRA measures to use and how to interpret it is contentious \cite{10.1007/978-3-319-77249-3_6, McHugh2012-iy}.  Using an inappropriate metric, such as Fleiss-$\kappa$ for interval data, is common in NLG \cite{amidei-etal-2019-agreement}. Furthermore,   even when an appropriate class of IRA metric is used, depending on the IRA chosen, the scores can range from poor to almost perfect \cite{10.1007/978-3-319-77249-3_6}. 

An intuitive measure of inter-rater agreement is using percent agreement. However, the main criticism was that this does not take chance agreement into account \cite{McHugh2012-iy}. The question to ask when using IRA metrics, such as a $kappa$ statistic is why and when does chance agreement matter? Some tasks genuinely have a class imbalance, e.g., span annotation tasks for named entity recognition, and unmarked spans for the majority class, which would lead to inflated percentage agreement \cite{Artstein2017}. Another reason is the assumption that some annotators might be making random guesses when they don't know  the answer, and that the majority of the raters may NOT be making deliberate choices \cite{McHugh2012-iy}. Hence, various IRA measures of Kappa (e.g., Cohen's-$\kappa$) or Alpha (e.g., Krippendorff's-$\alpha$), estimate the \textit{observed} chance agreement or disagreement empirically when computing IRA. However, if we consider a group of conscientious raters, does chance agreement matter for rating model outputs? For instance, if a model is a high performer and the most common rating is a 4, should disagreeing on a minority rating such as a 1 \textit{vs.} 2 drastically reduce the IRA? We further demonstrate, using a toy example in Appendix~\ref{app:sec:toyexample}, how disagreement on minority labels can substantially reduce IRA measured using a Krippendorff's-$\alpha$ while percentage agreement barely changes. 


What is considered as ``low'' IRA can vary from task to task, as complex tasks or difficult samples  tend to have low IRA \cite{kim2023interannotator}. This is crucially important for interpreting IRA and is particularly relevant for NLG evaluation. NLG tasks have typically reported relatively low IRA, e.g., average Krippendorff's-$\alpha$ of 0.62  \cite{amidei-etal-2019-agreement}, the standard interpretation is that experiments with scores less than 0.67 must be deemed unreliable \cite{MARZI2024102545}. For instance, rating ``\textit{How good is the generated story?}'' is more likely to have a lower IRA compared to ``\textit{Is 99 the largest 2-digit number}''.

Hence, while it is mandatory to measure IRA, it is important to ensure that the scores are interpreted in the context of the task. Performing detailed analysis including computing IRA using multiple metrics such as baseline percentage agreement and visualizing the item-wise agreement scores can help analyze the results. Researchers have also called for further investigation to understand the usefulness of a measure for a given problem, demonstrating the challenges in selecting an appropriate IRA measure \cite{10.1007/978-3-319-77249-3_6}.

\subsection{\underline{\textit{S}}coring cr\underline{\textit{i}}teria}
Scoring criteria refers to \textit{``What aspects to score"}. The 4 common scoring criteria covered in NLP literature are \textbf{(a)} Linguistic Fluency - the quality of single sentence \textbf{(b)} Coherence - overall flow or readability \textbf{(c)} Relevance - importance of content \textbf{(d)} Factuality - factual correctness \cite{10.1162/tacl_a_00373/Fabbri_summeval, gao-wan-2022-dialsummeval}. Also note that the nomenclature used to indicate a given criterion need not be consistent, e.g., as fluency vs. naturalness, across various studies \cite{van-der-lee-etal-2019-best}. Evaluation criteria also need to be customized across NLP tasks, and this is necessary as \textbf{(a)} the criteria will vary between domains and tasks \textbf{(b)} the generated text almost always needs to be evaluated against multiple criteria \cite{burchardt-2013-multidimensional,freitag-etal-2022-accelerating}. For instance, \citet{freitag-etal-2022-accelerating} propose the use of Multidimensional quality metrics (MQM) for evaluating machine translation. MQM is a generic framework for evaluating translation quality and provides a catalog of over 100 issues or error types organized in a hierarchy that evaluators can check for \cite{burchardt-2013-multidimensional}. This hierarchical categorization of errors enables granular as well as coarse-grained analysis of the quality of translation, and can be adapted for NLG evaluations.

In addition to the exhaustive evaluation categorization provided in the MQM framework, responsible AI (RAI) must be factored into human evaluation. \citet{sun2024trustllm} propose 6  categories for RAI evaluation -- truthfulness, safety, fairness, robustness, privacy, and machine ethics. Domain-specific customization and extensions also form an integral part of evaluation. For instance, evaluating the effectiveness of an LLM for a domain such as legal should include additional criteria such as case analysis and charge damages calculation \cite{fei2023lawbench}. See conceptual view in  Table~\ref{tab:scoringcriteria}.

\begin{table}[h!]
    \centering\small
    \begin{tabular}{ll}
    \toprule
        \textbf{High level category}& \textbf{Sub criteria} \\
    \midrule
         Core NLP & Fluency  \\
         Core NLP & Factuality  \\  
         Core NLP & Relevance  \\
         Core NLP & Coherence  \\
    \midrule
         Domain  Specific & ...  \\
    \midrule
        Responsible AI & Bias \& Fairness  \\
        Responsible AI & Privacy  \\
        Responsible AI & Safety  \\
        Responsible AI & Robustness  \\
        
    \bottomrule
    \end{tabular}
    \vspace{-1ex}
    \caption{Logical view  of high-level scoring criteria}
    \label{tab:scoringcriteria}
\vspace{-6mm}
\end{table}

\subsection{\textit{\underline{D}}ifferentiating} \label{sec:Differentiating}
``\textit{To differentiate is to identify the differences between things}'' \cite{vocabulary}. The test sets used in evaluation must be able to differentiate between the various capabilities as well as the weaknesses of generative LLMs. For instance, when the test sets are relatively easy, most models can achieve high scores and seem very capable. Conversely, when the test sets are very difficult, models might achieve low scores, making the models seem ineffective. Therefore, tests that are ineffective in differentiating capabilities can lead to \textbf{(a)} incorrect conclusions about the capabilities of a model, \textbf{(b)} poor calibration or ranking of various models, or \textbf{(c)} inability to identify any improvements or degradation between model versions, despite spending significant resources retraining new models. Hence, constructing effective tests that can differentiate model capabilities is crucial, otherwise evaluation simply results in wasted effort.

Traditionally, models have been evaluated using public datasets and benchmark such as GLUE~\citep{wang2019glue} and SuperGLUE~\citep{wang2019superglue}. These evaluations are not without their share of problems as models can exploit weaknesses in the official test sets relying on shortcuts or spurious correlation -- such as length of the input -- to predict the target label achieving high performance yet non-generalizable beyond the official test set \cite{mccoy-etal-2019-right, elangovan2023principles, gururangan-etal-2018-annotation}. In addition to these problems, evaluating the current generation of SOTA LLMs poses further challenges. \textbf{1)} LLMs are trained on billions of tokens available on the internet, and the training data used is rarely well documented. Hence, these LLMs may have consumed public benchmark test sets as part of their training data \cite{magar-schwartz-2022-data, sainz-etal-2023-nlp}. \textbf{2)} The models are generative, producing natural language output unlike a model trained on a classification task, making it difficult to automate evaluation and thus are far more reliant on human evaluation. \textbf{3)} The LLMs are capable of following natural language instructions to solve a vast variety of tasks, hence they need to be evaluated on a range of instructions provided as prompts that \ul{emulate end-user use cases}. 

Emulating end-user use cases is crucial, as there are fundamental differences in the tasks in NLP benchmarks vs. the kind of questions or problems  users can ask an LLM. Firstly, end users tend to be very creative and prompt the LLM for all kinds of queries. This diversity is particularly important in safety critical applications such as Medicine, as evaluation blindspots can  potentially lead to harmful consequences.    Secondly, the same message or question can be framed (prompted) in many ways. Hence, test scenarios need to consider the various natural ways in which users write their intentions when interacting with a LLM. If these aspects are not taken account when constructing a test set, solely relying on traditional NLP tasks  such as sentiment analysis  can result in traditional SOTA models outperforming ChatGPT \cite{KOCON2023101861}, when clearly this is not reflective of the end-user experience. Hence, we argue that the test cases used for evaluating LLMs need to represent end user scenarios, in addition to standard NLP tasks. Benchmarks such as Big-Bench \cite{srivastava2023beyond} take a step towards this direction by enabling GitHub contributors to add new tasks to the benchmark. Furthermore, the same prompts may not be effective against all LLMs. Hence,  prompts might often have to be customized for individual models, making benchmarking non-trivial. 

In light of these new challenges, \textbf{curating effective test sets is a critical problem} that the NLP community \textit{must} tackle. Benchmarks such as HELM \cite{liang2023holistic} to evaluate LLMs use multiple public datasets across various tasks such as Question Answering (QA) and sentiment analysis and measure aspects beyond accuracy to include toxicity and bias. While these are steps in the right direction, the effectiveness of public test sets such as IMDB dataset \cite{maas-EtAl:2011:ACL-HLT2011} for sentiment analysis or XSum for summarization \cite{narayan-etal-2018-dont} in measuring the capabilities of SOTA LLMs can be limited for the reasons discussed earlier. These datasets simply do not sufficiently represent end-user use cases. XSum also has hallucinated content in over 75\% of its gold summaries \cite{maynez-etal-2020-faithfulness}, demonstrating further weaknesses in the test sets themselves. 


Robustness testing is also a key aspect of evaluation, as models can be vulnerable to basic perturbations such as capitalization, white spaces and prompt formatting \cite{sclar2024quantifying} and need not understand basic linguistic concepts such as negation \citep{rogers-2021-changing}. While efforts such as DynaBench to curate progressively harder test sets \cite{kiela-etal-2021-dynabench} and behavior testing of models \cite{ribeiro-etal-2020-beyond} are promising approaches to curate effective test sets, curation still needs more research to make significant progress to target end-user use cases to evaluate LLMs.

\subsection{\textit{\underline{R}}esponsible}
In human evaluation, the Responsible pillar has to consider two aspects, \textbf{a)} Is the model behavior responsible? \textbf{b)} Do the human evaluators introduce bias to the evaluation results? While there is no formal definition of Responsible AI, at the very least it entails fairness, safety, truthfulness, and privacy \cite{sun2024trustllm}.
Searching papers with ``responsible'' in their title results in 20 papers, expanding the search to include terms such as ``bias'' and ``privacy'' results in $\approx$1000 or 1\% of the papers in ACL Anthology, while none of the human evaluation papers mention responsible AI related keywords in their title or abstract (query in Appendix~\ref{app:sec:responsibleAI}), a telltale sign that more work is needed.


\textbf{1. Is the model behavior responsible?} Answering this question requires evaluating the model for bias, its ability to withstand privacy attacks, truthfulness, and whether the responses are safe. \textbf{Bias} is ``\textit{prejudice in favor of or against one thing, person, or group}''. Bias affects groups by gender, race, culture, religion, geography, and disability \cite{esiobu-etal-2023-robbie}. Thus, the tests sent for human evaluation need to specifically cater to these cases and report the performance of various subgroups to ensure that the LLM behaves responsibly. This requires that the tests have metadata curated such as race to be able to report across these segments. 

Generative LLMs are susceptible to leaking private details such as person identifiable information (PII) from the training data, including when subject to \textbf{privacy} attacks \cite{vakili-dalianis-2023-using, li-etal-2023-multi-step}. One method to mitigate such privacy leaks is to obfuscate PII information such as names and locations from training data using techniques such as Pseudonymization -- recognizing privacy-sensitive information and replacing them with realistic substitute \cite{vakili-dalianis-2023-using} and differential-privacy-based approaches that add noise to the input \cite{chen-etal-2023-customized}. \textbf{Safety} ensures that the models do not generate content that harms a person's physical safety or mental health \cite{mei-etal-2023-assert, rusert-etal-2022-robustness}. Red teams, a group of people authorized to imitate an adversary's attack or exploitation capabilities, are  used to evaluate the robustness, safety, and privacy of LLMs \cite{perez-etal-2022-red, radharapu-etal-2023-aart}.

\textbf{2. Do the human evaluators introduce bias to the evaluation results?} As discussed previously, when a substantial portion of evaluation relies on human perception, the diversity of the human evaluators plays a key role in how representative the results are of the wider population. Despite this, less than 3\% of papers report demographic information about their evaluators \cite{VANDERLEE2021101151}. Having a small group of evaluators or even when the size is large, aspects such as selection bias can result in biased results. Hence, we call for human evaluation to consider the evaluator demographic to mitigate bias effects in the evaluation.

\subsection{\underline{\textit{S}}calability}
Human evaluation is expensive, yet large volumes of test cases are necessary to differentiate model capabilities and ensure consistency. Hence,  optimizations to reduce cost and time is crucial for wider adoption. Automating  parts of human evaluation can reduce cost. For instance, automation might be potentially helpful to shortlist a set of candidates for human evaluation. 
While the shortcomings of n-gram-based automated evaluations using metrics like Rouge \cite{lin-2004-rouge} are well studied \cite{deutsch-etal-2022-examining}, approaches such as using LLMs to evaluate LLMs \cite{lin-chen-2023-llm, chiang-lee-2023-closer} need further exploration. Firstly, the effectiveness of LLM-based evaluation is measured using its correlation with human evaluation, hence the human evaluation procedures need to be strengthened first to draw comprehensive and robust conclusions, creating a chicken-and-egg problem. Secondly, LLMs to evaluate LLMs should take into account the differences between perception-based metrics and evaluations that rely on facts, as discussed in Section~\ref{sec:mitigatecognitivebias}.

Studies that attempt to reduce the turn-around times of human evaluation itself are limited, less than 50 papers mention ``cost'' or ``scale'' with human eval in the title (see Appendix~\ref{app:sec:humanevalscalecost} for search query) in the last 20 years. \citet{levinboim-etal-2021-quality} report using coarse-grained caption (as opposed to fine-grained) annotations from crowdsourced users to be able to scale, a method that can be adopted for human evaluation. \citet{huang-etal-2023-reduce} propose to identify effective test samples to reduce the cost of human evaluation in conversation systems. Designing a UI that enables the human annotators to work efficiently (one of the Es Efficiency in usability in section~\ref{sec:ExperimentalDesign}) can reduce time. Given the limited amount of work in this area of scalability of human evaluation, we call for further research.

\section{Conclusion \& Moving forward}
\vspace{-1mm}
We presented the \textit{ConSiDERS-The-Human} evaluation framework to keep up with the increasing capabilities of LLMs. We highlight the effects of human emotions and cognitive biases on evaluation, given how commonly the perception-based metrics are used to evaluate aspects such as truthfulness. We, hence, encourage researchers in NLP to collaborate with their counterparts in UX, HCI \& psychology to ensure that the evaluation \textit{\ul{measures the right things the right way}} and the results are interpreted accurately. We also call for further research in critical areas -- including curating effective test sets, scalability of human evaluation, and responsible AI components such as privacy, bias, robustness, and safety considerations, when evaluating increasingly powerful \& ubiquitous generative LLMs. 

\section{Limitations}
Human evaluation is a challenge, especially given the increasing capabilities of SOTA LLMs. Firstly, LLMs have many potential applications and effectively evaluating each application or domain might need customization. Our main aim in this paper is to provide a generic framework extensible for specific domains or applications. Effectively customizing for individual cases might require trial and error. Secondly, perception, whilst important, cannot solely dictate how the quality of LLMs is measured. Humans use heuristics to make decisions, and evaluation has to cater to these heuristics. While there are over 180 cognitive biases, in our paper we only highlight a few that can impact  evaluation, we specify this limitation in Section~\ref{sec:cognitivebias} as well.   

\section*{Acknowledgements}
We would like to thank our colleagues Seonwoo Min and Han-Chin Shing for reviewing this paper and providing feedback.

\bibliography{anthology,custom}

\begin{thebibliography}{113}
\expandafter\ifx\csname natexlab\endcsname\relax\def\natexlab#1{#1}\fi

\bibitem[{Agley et~al.(2022)Agley, Xiao, Nolan, and Golzarri-Arroyo}]{Agley2022}
Jon Agley, Yunyu Xiao, Rachael Nolan, and Lilian Golzarri-Arroyo. 2022.
\newblock \href {https://doi.org/10.3758/s13428-021-01665-8} {Quality control questions on amazon's mechanical turk (mturk): A randomized trial of impact on the usaudit, phq-9, and gad-7}.
\newblock \emph{Behavior Research Methods}, 54(2):885--897.

\bibitem[{Alagarai~Sampath et~al.(2014)Alagarai~Sampath, Rajeshuni, and Indurkhya}]{Sampath10.1145/2556288.2557155}
Harini Alagarai~Sampath, Rajeev Rajeshuni, and Bipin Indurkhya. 2014.
\newblock \href {https://doi.org/10.1145/2556288.2557155} {Cognitively inspired task design to improve user performance on crowdsourcing platforms}.
\newblock In \emph{Proceedings of the SIGCHI Conference on Human Factors in Computing Systems}, CHI '14, page 3665–3674, New York, NY, USA. Association for Computing Machinery.

\bibitem[{Amatriain et~al.(2009{\natexlab{a}})Amatriain, Pujol, and Oliver}]{10.1007/978-3-642-02247-0_24}
Xavier Amatriain, Josep~M. Pujol, and Nuria Oliver. 2009{\natexlab{a}}.
\newblock I like it... i like it not: Evaluating user ratings noise in recommender systems.
\newblock In \emph{User Modeling, Adaptation, and Personalization}, pages 247--258, Berlin, Heidelberg. Springer Berlin Heidelberg.

\bibitem[{Amatriain et~al.(2009{\natexlab{b}})Amatriain, Pujol, Tintarev, and Oliver}]{10.1145/1639714.1639744}
Xavier Amatriain, Josep~M. Pujol, Nava Tintarev, and Nuria Oliver. 2009{\natexlab{b}}.
\newblock \href {https://doi.org/10.1145/1639714.1639744} {Rate it again: increasing recommendation accuracy by user re-rating}.
\newblock In \emph{Proceedings of the Third ACM Conference on Recommender Systems}, RecSys '09, page 173–180, New York, NY, USA. Association for Computing Machinery.

\bibitem[{Amidei et~al.(2019)Amidei, Piwek, and Willis}]{amidei-etal-2019-agreement}
Jacopo Amidei, Paul Piwek, and Alistair Willis. 2019.
\newblock \href {https://doi.org/10.18653/v1/W19-8642} {Agreement is overrated: A plea for correlation to assess human evaluation reliability}.
\newblock In \emph{Proceedings of the 12th International Conference on Natural Language Generation}, pages 344--354, Tokyo, Japan. Association for Computational Linguistics.

\bibitem[{Andreas~Sonderegger and Sauer(2012)}]{doi:10.1080/00140139.2012.672658}
Andreas~Uebelbacher Andreas~Sonderegger, Gerold~Zbinden and Juergen Sauer. 2012.
\newblock \href {https://doi.org/10.1080/00140139.2012.672658} {The influence of product aesthetics and usability over the course of time: a longitudinal field experiment}.
\newblock \emph{Ergonomics}, 55(7):713--730.
\newblock PMID: 22506866.

\bibitem[{Artstein(2017)}]{Artstein2017}
Ron Artstein. 2017.
\newblock \href {https://doi.org/10.1007/978-94-024-0881-2_11} {\emph{Inter-annotator Agreement}}, pages 297--313. Springer Netherlands, Dordrecht.

\bibitem[{Azzopardi(2021)}]{10.1145/3406522.3446023}
Leif Azzopardi. 2021.
\newblock \href {https://doi.org/10.1145/3406522.3446023} {Cognitive biases in search: A review and reflection of cognitive biases in information retrieval}.
\newblock In \emph{Proceedings of the 2021 Conference on Human Information Interaction and Retrieval}, CHIIR '21, page 27–37, New York, NY, USA. Association for Computing Machinery.

\bibitem[{Bailey(1993)}]{doi:10.1177/154193129303700406/RobertWBailey}
Robert~W. Bailey. 1993.
\newblock \href {https://doi.org/10.1177/154193129303700406} {Performance vs. preference}.
\newblock \emph{Proceedings of the Human Factors and Ergonomics Society Annual Meeting}, 37(4):282--286.

\bibitem[{Baker(2016)}]{Baker2016}
Monya Baker. 2016.
\newblock \href {https://doi.org/10.1038/533452a} {1,500 scientists lift the lid on reproducibility}.
\newblock \emph{Nature}, 533(7604):452--454.

\bibitem[{Barnum(2020)}]{barnum2020usability}
Carol~M Barnum. 2020.
\newblock \emph{Usability testing essentials: Ready, set... test!}
\newblock Morgan Kaufmann.

\bibitem[{Belz and Kow(2010)}]{belz-kow-2010-comparing}
Anja Belz and Eric Kow. 2010.
\newblock \href {https://aclanthology.org/W10-4201} {Comparing rating scales and preference judgements in language evaluation}.
\newblock In \emph{Proceedings of the 6th International Natural Language Generation Conference}. Association for Computational Linguistics.

\bibitem[{Belz et~al.(2020)Belz, Mille, and Howcroft}]{belz-etal-2020-disentangling}
Anya Belz, Simon Mille, and David~M. Howcroft. 2020.
\newblock \href {https://aclanthology.org/2020.inlg-1.24} {Disentangling the properties of human evaluation methods: A classification system to support comparability, meta-evaluation and reproducibility testing}.
\newblock In \emph{Proceedings of the 13th International Conference on Natural Language Generation}, pages 183--194, Dublin, Ireland. Association for Computational Linguistics.

\bibitem[{Belz et~al.(2023)Belz, Thomson, Reiter, and Mille}]{belz-etal-2023-non}
Anya Belz, Craig Thomson, Ehud Reiter, and Simon Mille. 2023.
\newblock \href {https://doi.org/10.18653/v1/2023.findings-acl.226} {Non-repeatable experiments and non-reproducible results: The reproducibility crisis in human evaluation in {NLP}}.
\newblock In \emph{Findings of the Association for Computational Linguistics: ACL 2023}, pages 3676--3687, Toronto, Canada. Association for Computational Linguistics.

\bibitem[{bench authors(2023)}]{srivastava2023beyond}
Big bench authors. 2023.
\newblock \href {https://openreview.net/forum?id=uyTL5Bvosj} {Beyond the imitation game: Quantifying and extrapolating the capabilities of language models}.
\newblock \emph{Transactions on Machine Learning Research}.

\bibitem[{Bishop and Herron(2015)}]{Bishop2015-dk}
Phillip~A Bishop and Robert~L Herron. 2015.
\newblock Use and misuse of the likert item responses and other ordinal measures.
\newblock \emph{Int. J. Exerc. Sci.}, 8(3):297--302.

\bibitem[{Brashier and Marsh(2020)}]{doi:10.1146/annurev-psych-010419-050807}
Nadia~M. Brashier and Elizabeth~J. Marsh. 2020.
\newblock \href {https://doi.org/10.1146/annurev-psych-010419-050807} {Judging truth}.
\newblock \emph{Annual Review of Psychology}, 71(1):499--515.
\newblock PMID: 31514579.

\bibitem[{Burchardt(2013)}]{burchardt-2013-multidimensional}
Aljoscha Burchardt. 2013.
\newblock \href {https://aclanthology.org/2013.tc-1.6} {Multidimensional quality metrics: a flexible system for assessing translation quality}.
\newblock In \emph{Proceedings of Translating and the Computer 35}, London, UK. Aslib.

\bibitem[{Button et~al.(2013)Button, Ioannidis, Mokrysz, Nosek, Flint, Robinson, and Munaf{\`o}}]{Button2013}
Katherine~S. Button, John P.~A. Ioannidis, Claire Mokrysz, Brian~A. Nosek, Jonathan Flint, Emma S.~J. Robinson, and Marcus~R. Munaf{\`o}. 2013.
\newblock \href {https://doi.org/10.1038/nrn3475} {Power failure: why small sample size undermines the reliability of neuroscience}.
\newblock \emph{Nature Reviews Neuroscience}, 14(5):365--376.

\bibitem[{Chen et~al.(2023)Chen, Mo, Wang, Chen, Nie, Wang, and Cui}]{chen-etal-2023-customized}
Sai Chen, Fengran Mo, Yanhao Wang, Cen Chen, Jian-Yun Nie, Chengyu Wang, and Jamie Cui. 2023.
\newblock \href {https://doi.org/10.18653/v1/2023.findings-acl.355} {A customized text sanitization mechanism with differential privacy}.
\newblock In \emph{Findings of the Association for Computational Linguistics: ACL 2023}, pages 5747--5758, Toronto, Canada. Association for Computational Linguistics.

\bibitem[{Chiang and Lee(2023)}]{chiang-lee-2023-closer}
Cheng-Han Chiang and Hung-yi Lee. 2023.
\newblock \href {https://doi.org/10.18653/v1/2023.findings-emnlp.599} {A closer look into using large language models for automatic evaluation}.
\newblock In \emph{Findings of the Association for Computational Linguistics: EMNLP 2023}, pages 8928--8942, Singapore. Association for Computational Linguistics.

\bibitem[{Choi and Pak(2005)}]{Choi2005-rs}
Bernard C~K Choi and Anita W~P Pak. 2005.
\newblock A catalog of biases in questionnaires.
\newblock \emph{Prev. Chronic Dis.}, 2(1):A13.

\bibitem[{Clark et~al.(2021)Clark, August, Serrano, Haduong, Gururangan, and Smith}]{clark-etal-2021-thats}
Elizabeth Clark, Tal August, Sofia Serrano, Nikita Haduong, Suchin Gururangan, and Noah~A. Smith. 2021.
\newblock \href {https://doi.org/10.18653/v1/2021.acl-long.565} {All that{'}s {`}human{'} is not gold: Evaluating human evaluation of generated text}.
\newblock In \emph{Proceedings of the 59th Annual Meeting of the Association for Computational Linguistics and the 11th International Joint Conference on Natural Language Processing (Volume 1: Long Papers)}, pages 7282--7296, Online. Association for Computational Linguistics.

\bibitem[{Deutsch et~al.(2022)Deutsch, Dror, and Roth}]{deutsch-etal-2022-examining}
Daniel Deutsch, Rotem Dror, and Dan Roth. 2022.
\newblock \href {https://doi.org/10.18653/v1/2022.naacl-main.442} {Re-examining system-level correlations of automatic summarization evaluation metrics}.
\newblock In \emph{Proceedings of the 2022 Conference of the North American Chapter of the Association for Computational Linguistics: Human Language Technologies}, pages 6038--6052, Seattle, United States. Association for Computational Linguistics.

\bibitem[{Eickhoff(2018)}]{10.1145/3159652.3159654}
Carsten Eickhoff. 2018.
\newblock \href {https://doi.org/10.1145/3159652.3159654} {Cognitive biases in crowdsourcing}.
\newblock In \emph{Proceedings of the Eleventh ACM International Conference on Web Search and Data Mining}, WSDM '18, page 162–170, New York, NY, USA. Association for Computing Machinery.

\bibitem[{Elangovan et~al.(2023)Elangovan, He, Li, and Verspoor}]{elangovan2023principles}
Aparna Elangovan, Jiayuan He, Yuan Li, and Karin Verspoor. 2023.
\newblock \href {http://arxiv.org/abs/2311.03663} {Principles from clinical research for nlp model generalization}.

\bibitem[{Elangovan et~al.(2021)Elangovan, He, and Verspoor}]{elangovan-etal-2021-memorization}
Aparna Elangovan, Jiayuan He, and Karin Verspoor. 2021.
\newblock \href {https://doi.org/10.18653/v1/2021.eacl-main.113} {Memorization vs. generalization : Quantifying data leakage in {NLP} performance evaluation}.
\newblock In \emph{Proceedings of the 16th Conference of the European Chapter of the Association for Computational Linguistics: Main Volume}, pages 1325--1335, Online. Association for Computational Linguistics.

\bibitem[{Ellis(2018)}]{Ellis2018}
Geoffrey Ellis. 2018.
\newblock \href {https://doi.org/10.1007/978-3-319-95831-6_1} {\emph{So, What Are Cognitive Biases?}}, pages 1--10. Springer International Publishing, Cham.

\bibitem[{Esiobu et~al.(2023)Esiobu, Tan, Hosseini, Ung, Zhang, Fernandes, Dwivedi-Yu, Presani, Williams, and Smith}]{esiobu-etal-2023-robbie}
David Esiobu, Xiaoqing Tan, Saghar Hosseini, Megan Ung, Yuchen Zhang, Jude Fernandes, Jane Dwivedi-Yu, Eleonora Presani, Adina Williams, and Eric Smith. 2023.
\newblock \href {https://doi.org/10.18653/v1/2023.emnlp-main.230} {{ROBBIE}: Robust bias evaluation of large generative language models}.
\newblock In \emph{Proceedings of the 2023 Conference on Empirical Methods in Natural Language Processing}, pages 3764--3814, Singapore. Association for Computational Linguistics.

\bibitem[{Fabbri et~al.(2021)Fabbri, Kryściński, McCann, Xiong, Socher, and Radev}]{10.1162/tacl_a_00373/Fabbri_summeval}
Alexander~R. Fabbri, Wojciech Kryściński, Bryan McCann, Caiming Xiong, Richard Socher, and Dragomir Radev. 2021.
\newblock \href {https://doi.org/10.1162/tacl_a_00373} {{SummEval: Re-evaluating Summarization Evaluation}}.
\newblock \emph{Transactions of the Association for Computational Linguistics}, 9:391--409.

\bibitem[{Fei et~al.(2023)Fei, Shen, Zhu, Zhou, Han, Zhang, Chen, Shen, and Ge}]{fei2023lawbench}
Zhiwei Fei, Xiaoyu Shen, Dawei Zhu, Fengzhe Zhou, Zhuo Han, Songyang Zhang, Kai Chen, Zongwen Shen, and Jidong Ge. 2023.
\newblock \href {http://arxiv.org/abs/2309.16289} {Lawbench: Benchmarking legal knowledge of large language models}.

\bibitem[{Freitag et~al.(2022)Freitag, Cadigan, Niekrasz, and Sasseen}]{freitag-etal-2022-accelerating}
Dayne Freitag, John Cadigan, John Niekrasz, and Robert Sasseen. 2022.
\newblock \href {https://aclanthology.org/2022.pandl-1.6} {Accelerating human authorship of information extraction rules}.
\newblock In \emph{Proceedings of the First Workshop on Pattern-based Approaches to NLP in the Age of Deep Learning}, pages 45--55, Gyeongju, Republic of Korea. International Conference on Computational Linguistics.

\bibitem[{Freitag et~al.(2021)Freitag, Foster, Grangier, Ratnakar, Tan, and Macherey}]{freitag-etal-2021-experts}
Markus Freitag, George Foster, David Grangier, Viresh Ratnakar, Qijun Tan, and Wolfgang Macherey. 2021.
\newblock \href {https://doi.org/10.1162/tacl_a_00437} {Experts, errors, and context: A large-scale study of human evaluation for machine translation}.
\newblock \emph{Transactions of the Association for Computational Linguistics}, 9:1460--1474.

\bibitem[{Furnham and Boo(2011)}]{FURNHAM201135}
Adrian Furnham and Hua~Chu Boo. 2011.
\newblock \href {https://doi.org/https://doi.org/10.1016/j.socec.2010.10.008} {A literature review of the anchoring effect}.
\newblock \emph{The Journal of Socio-Economics}, 40(1):35--42.

\bibitem[{Gadiraju et~al.(2017)Gadiraju, Yang, and Bozzon}]{Gadiraju_10.1145/3078714.3078715}
Ujwal Gadiraju, Jie Yang, and Alessandro Bozzon. 2017.
\newblock \href {https://doi.org/10.1145/3078714.3078715} {Clarity is a worthwhile quality: On the role of task clarity in microtask crowdsourcing}.
\newblock In \emph{Proceedings of the 28th ACM Conference on Hypertext and Social Media}, HT '17, page 5–14, New York, NY, USA. Association for Computing Machinery.

\bibitem[{Gao and Wan(2022)}]{gao-wan-2022-dialsummeval}
Mingqi Gao and Xiaojun Wan. 2022.
\newblock \href {https://doi.org/10.18653/v1/2022.naacl-main.418} {{D}ial{S}umm{E}val: Revisiting summarization evaluation for dialogues}.
\newblock In \emph{Proceedings of the 2022 Conference of the North American Chapter of the Association for Computational Linguistics: Human Language Technologies}, pages 5693--5709, Seattle, United States. Association for Computational Linguistics.

\bibitem[{Gisev et~al.(2013)Gisev, Bell, and Chen}]{GISEV2013330}
Natasa Gisev, J.~Simon Bell, and Timothy~F. Chen. 2013.
\newblock \href {https://doi.org/https://doi.org/10.1016/j.sapharm.2012.04.004} {Interrater agreement and interrater reliability: Key concepts, approaches, and applications}.
\newblock \emph{Research in Social and Administrative Pharmacy}, 9(3):330--338.

\bibitem[{Greifeneder and Bless(2017)}]{greifeneder2017interplay}
Rainer Greifeneder and Herbert Bless. 2017.
\newblock The interplay of cognition and feelings: Fluency.
\newblock \emph{Social cognition}, pages 145--164.

\bibitem[{Gururangan et~al.(2018)Gururangan, Swayamdipta, Levy, Schwartz, Bowman, and Smith}]{gururangan-etal-2018-annotation}
Suchin Gururangan, Swabha Swayamdipta, Omer Levy, Roy Schwartz, Samuel Bowman, and Noah~A. Smith. 2018.
\newblock \href {https://doi.org/10.18653/v1/N18-2017} {Annotation artifacts in natural language inference data}.
\newblock In \emph{Proceedings of the 2018 Conference of the North {A}merican Chapter of the Association for Computational Linguistics: Human Language Technologies, Volume 2 (Short Papers)}, pages 107--112, New Orleans, Louisiana. Association for Computational Linguistics.

\bibitem[{H{\"a}m{\"a}l{\"a}inen and Alnajjar(2021)}]{hamalainen-alnajjar-2021-great}
Mika H{\"a}m{\"a}l{\"a}inen and Khalid Alnajjar. 2021.
\newblock \href {https://aclanthology.org/2021.humeval-1.8} {The great misalignment problem in human evaluation of {NLP} methods}.
\newblock In \emph{Proceedings of the Workshop on Human Evaluation of NLP Systems (HumEval)}, pages 69--74, Online. Association for Computational Linguistics.

\bibitem[{Hamborg et~al.(2014)Hamborg, H{\"u}lsmann, and Kaspar}]{Hamborg2014}
Kai-Christoph Hamborg, Julia H{\"u}lsmann, and Kai Kaspar. 2014.
\newblock \href {https://doi.org/10.1155/2014/946239} {The interplay between usability and aesthetics: More evidence for the ``what is usable is beautiful'' notion}.
\newblock \emph{Advances in Human-Computer Interaction}, 2014:946239.

\bibitem[{Hartson and Pyla(2012)}]{hartson2012ux}
Rex Hartson and Pardha~S Pyla. 2012.
\newblock \emph{The UX Book: Process and guidelines for ensuring a quality user experience}.
\newblock Elsevier.

\bibitem[{Hayhoe(1990)}]{HAYHOE1990677}
Douglas Hayhoe. 1990.
\newblock \href {https://doi.org/https://doi.org/10.1016/S0020-7373(05)80069-0} {Sorting-based menu categories}.
\newblock \emph{International Journal of Man-Machine Studies}, 33(6):677--705.

\bibitem[{Hill et~al.(1995)Hill, Stead, Rosenstein, and Furnas}]{hill1995recommending}
Will Hill, Larry Stead, Mark Rosenstein, and George Furnas. 1995.
\newblock Recommending and evaluating choices in a virtual community of use.
\newblock In \emph{Proceedings of the SIGCHI conference on Human factors in computing systems}, pages 194--201.

\bibitem[{Huang et~al.(2023)Huang, Qin, Lei, and Lv}]{huang-etal-2023-reduce}
Chen Huang, Peixin Qin, Wenqiang Lei, and Jiancheng Lv. 2023.
\newblock \href {https://doi.org/10.18653/v1/2023.emnlp-main.670} {Reduce human labor on evaluating conversational information retrieval system: A human-machine collaboration approach}.
\newblock In \emph{Proceedings of the 2023 Conference on Empirical Methods in Natural Language Processing}, pages 10876--10891, Singapore. Association for Computational Linguistics.

\bibitem[{Iskender et~al.(2021)Iskender, Polzehl, and M{\"o}ller}]{iskender-etal-2021-reliability}
Neslihan Iskender, Tim Polzehl, and Sebastian M{\"o}ller. 2021.
\newblock \href {https://aclanthology.org/2021.humeval-1.10} {Reliability of human evaluation for text summarization: Lessons learned and challenges ahead}.
\newblock In \emph{Proceedings of the Workshop on Human Evaluation of NLP Systems (HumEval)}, pages 86--96, Online. Association for Computational Linguistics.

\bibitem[{Jasberg and Sizov(2020)}]{doi:10.1080/0144929X.2019.1604804}
Kevin Jasberg and Sergej Sizov. 2020.
\newblock \href {https://doi.org/10.1080/0144929X.2019.1604804} {Human uncertainty in explicit user feedback and its impact on the comparative evaluations of accurate prediction and personalisation}.
\newblock \emph{Behaviour \& Information Technology}, 39(5):544--577.

\bibitem[{Joorabloo et~al.(2022)Joorabloo, Jalili, and Ren}]{JOORABLOO2022242}
Nima Joorabloo, Mahdi Jalili, and Yongli Ren. 2022.
\newblock \href {https://doi.org/https://doi.org/10.1016/j.ins.2022.03.068} {Improved recommender systems by denoising ratings in highly sparse datasets through individual rating confidence}.
\newblock \emph{Information Sciences}, 601:242--254.

\bibitem[{Kaplan et~al.(2014)Kaplan, Chambers, and Glasgow}]{Kaplan2014-dq}
Robert~M Kaplan, David~A Chambers, and Russell~E Glasgow. 2014.
\newblock Big data and large sample size: a cautionary note on the potential for bias.
\newblock \emph{Clin. Transl. Sci.}, 7(4):342--346.

\bibitem[{Kiela et~al.(2021)Kiela, Bartolo, Nie, Kaushik, Geiger, Wu, Vidgen, Prasad, Singh, Ringshia, Ma, Thrush, Riedel, Waseem, Stenetorp, Jia, Bansal, Potts, and Williams}]{kiela-etal-2021-dynabench}
Douwe Kiela, Max Bartolo, Yixin Nie, Divyansh Kaushik, Atticus Geiger, Zhengxuan Wu, Bertie Vidgen, Grusha Prasad, Amanpreet Singh, Pratik Ringshia, Zhiyi Ma, Tristan Thrush, Sebastian Riedel, Zeerak Waseem, Pontus Stenetorp, Robin Jia, Mohit Bansal, Christopher Potts, and Adina Williams. 2021.
\newblock \href {https://doi.org/10.18653/v1/2021.naacl-main.324} {Dynabench: Rethinking benchmarking in {NLP}}.
\newblock In \emph{Proceedings of the 2021 Conference of the North American Chapter of the Association for Computational Linguistics: Human Language Technologies}, pages 4110--4124, Online. Association for Computational Linguistics.

\bibitem[{Kim and Park(2023)}]{kim2023interannotator}
NamHyeok Kim and Chanjun Park. 2023.
\newblock \href {http://arxiv.org/abs/2306.14373} {Inter-annotator agreement in the wild: Uncovering its emerging roles and considerations in real-world scenarios}.

\bibitem[{Koch and Forgas(2012)}]{KOCH2012481}
Alex~S. Koch and Joseph~P. Forgas. 2012.
\newblock \href {https://doi.org/https://doi.org/10.1016/j.jesp.2011.10.006} {Feeling good and feeling truth: The interactive effects of mood and processing fluency on truth judgments}.
\newblock \emph{Journal of Experimental Social Psychology}, 48(2):481--485.

\bibitem[{Kocoń et~al.(2023)Kocoń, Cichecki, Kaszyca, Kochanek, Szydło, Baran, Bielaniewicz, Gruza, Janz, Kanclerz, Kocoń, Koptyra, Mieleszczenko-Kowszewicz, Miłkowski, Oleksy, Piasecki, Łukasz Radliński, Wojtasik, Woźniak, and Kazienko}]{KOCON2023101861}
Jan Kocoń, Igor Cichecki, Oliwier Kaszyca, Mateusz Kochanek, Dominika Szydło, Joanna Baran, Julita Bielaniewicz, Marcin Gruza, Arkadiusz Janz, Kamil Kanclerz, Anna Kocoń, Bartłomiej Koptyra, Wiktoria Mieleszczenko-Kowszewicz, Piotr Miłkowski, Marcin Oleksy, Maciej Piasecki, Łukasz Radliński, Konrad Wojtasik, Stanisław Woźniak, and Przemysław Kazienko. 2023.
\newblock \href {https://doi.org/https://doi.org/10.1016/j.inffus.2023.101861} {Chatgpt: Jack of all trades, master of none}.
\newblock \emph{Information Fusion}, 99:101861.

\bibitem[{Kotkov et~al.(2022)Kotkov, Medlar, Satyal, Maslov, Neovius, and Glowacka}]{Kotkov2022RatingCI}
Denis Kotkov, Alan Medlar, Umesh~Raj Satyal, Alexandr~V. Maslov, Mats Neovius, and Dorota Glowacka. 2022.
\newblock \href {https://api.semanticscholar.org/CorpusID:248545996} {Rating consistency is consistently underrated: an exploratory analysis of movie-tag rating inconsistency}.
\newblock \emph{Proceedings of the 37th ACM/SIGAPP Symposium on Applied Computing}.

\bibitem[{Lau and Coiera(2007)}]{10.1197/jamia.M2411}
Annie~Y.S. Lau and Enrico~W. Coiera. 2007.
\newblock \href {https://doi.org/10.1197/jamia.M2411} {{Do People Experience Cognitive Biases while Searching for Information?}}
\newblock \emph{Journal of the American Medical Informatics Association}, 14(5):599--608.

\bibitem[{Leung(2011)}]{doi:10.1080/01488376.2011.580697}
Shing-On Leung. 2011.
\newblock \href {https://doi.org/10.1080/01488376.2011.580697} {A comparison of psychometric properties and normality in 4-, 5-, 6-, and 11-point likert scales}.
\newblock \emph{Journal of Social Service Research}, 37(4):412--421.

\bibitem[{Levinboim et~al.(2021)Levinboim, Thapliyal, Sharma, and Soricut}]{levinboim-etal-2021-quality}
Tomer Levinboim, Ashish~V. Thapliyal, Piyush Sharma, and Radu Soricut. 2021.
\newblock \href {https://doi.org/10.18653/v1/2021.naacl-main.253} {Quality estimation for image captions based on large-scale human evaluations}.
\newblock In \emph{Proceedings of the 2021 Conference of the North American Chapter of the Association for Computational Linguistics: Human Language Technologies}, pages 3157--3166, Online. Association for Computational Linguistics.

\bibitem[{Li et~al.(2023)Li, Guo, Fan, Xu, Huang, Meng, and Song}]{li-etal-2023-multi-step}
Haoran Li, Dadi Guo, Wei Fan, Mingshi Xu, Jie Huang, Fanpu Meng, and Yangqiu Song. 2023.
\newblock \href {https://doi.org/10.18653/v1/2023.findings-emnlp.272} {Multi-step jailbreaking privacy attacks on {C}hat{GPT}}.
\newblock In \emph{Findings of the Association for Computational Linguistics: EMNLP 2023}, pages 4138--4153, Singapore. Association for Computational Linguistics.

\bibitem[{Liang et~al.(2023)Liang, Bommasani, Lee, Tsipras, Soylu, Yasunaga, Zhang, Narayanan, Wu, Kumar, Newman, Yuan, Yan, Zhang, Cosgrove, Manning, Re, Acosta-Navas, Hudson, Zelikman, Durmus, Ladhak, Rong, Ren, Yao, WANG, Santhanam, Orr, Zheng, Yuksekgonul, Suzgun, Kim, Guha, Chatterji, Khattab, Henderson, Huang, Chi, Xie, Santurkar, Ganguli, Hashimoto, Icard, Zhang, Chaudhary, Wang, Li, Mai, Zhang, and Koreeda}]{liang2023holistic}
Percy Liang, Rishi Bommasani, Tony Lee, Dimitris Tsipras, Dilara Soylu, Michihiro Yasunaga, Yian Zhang, Deepak Narayanan, Yuhuai Wu, Ananya Kumar, Benjamin Newman, Binhang Yuan, Bobby Yan, Ce~Zhang, Christian~Alexander Cosgrove, Christopher~D Manning, Christopher Re, Diana Acosta-Navas, Drew~Arad Hudson, Eric Zelikman, Esin Durmus, Faisal Ladhak, Frieda Rong, Hongyu Ren, Huaxiu Yao, Jue WANG, Keshav Santhanam, Laurel Orr, Lucia Zheng, Mert Yuksekgonul, Mirac Suzgun, Nathan Kim, Neel Guha, Niladri~S. Chatterji, Omar Khattab, Peter Henderson, Qian Huang, Ryan~Andrew Chi, Sang~Michael Xie, Shibani Santurkar, Surya Ganguli, Tatsunori Hashimoto, Thomas Icard, Tianyi Zhang, Vishrav Chaudhary, William Wang, Xuechen Li, Yifan Mai, Yuhui Zhang, and Yuta Koreeda. 2023.
\newblock \href {https://openreview.net/forum?id=iO4LZibEqW} {Holistic evaluation of language models}.
\newblock \emph{Transactions on Machine Learning Research}.
\newblock Featured Certification, Expert Certification.

\bibitem[{Liebling et~al.(2022)Liebling, Heller, Robertson, and Deng}]{liebling-etal-2022-opportunities}
Daniel Liebling, Katherine Heller, Samantha Robertson, and Wesley Deng. 2022.
\newblock \href {https://doi.org/10.18653/v1/2022.findings-naacl.17} {Opportunities for human-centered evaluation of machine translation systems}.
\newblock In \emph{Findings of the Association for Computational Linguistics: NAACL 2022}, pages 229--240, Seattle, United States. Association for Computational Linguistics.

\bibitem[{Lin(2004)}]{lin-2004-rouge}
Chin-Yew Lin. 2004.
\newblock \href {https://aclanthology.org/W04-1013} {{ROUGE}: A package for automatic evaluation of summaries}.
\newblock In \emph{Text Summarization Branches Out}, pages 74--81, Barcelona, Spain. Association for Computational Linguistics.

\bibitem[{Lin and Chen(2023)}]{lin-chen-2023-llm}
Yen-Ting Lin and Yun-Nung Chen. 2023.
\newblock \href {https://doi.org/10.18653/v1/2023.nlp4convai-1.5} {{LLM}-eval: Unified multi-dimensional automatic evaluation for open-domain conversations with large language models}.
\newblock In \emph{Proceedings of the 5th Workshop on NLP for Conversational AI (NLP4ConvAI 2023)}, pages 47--58, Toronto, Canada. Association for Computational Linguistics.

\bibitem[{Liu et~al.(2023)Liu, Fabbri, Liu, Zhao, Nan, Han, Han, Joty, Wu, Xiong, and Radev}]{liu-etal-2023-revisiting}
Yixin Liu, Alex Fabbri, Pengfei Liu, Yilun Zhao, Linyong Nan, Ruilin Han, Simeng Han, Shafiq Joty, Chien-Sheng Wu, Caiming Xiong, and Dragomir Radev. 2023.
\newblock \href {https://doi.org/10.18653/v1/2023.acl-long.228} {Revisiting the gold standard: Grounding summarization evaluation with robust human evaluation}.
\newblock In \emph{Proceedings of the 61st Annual Meeting of the Association for Computational Linguistics (Volume 1: Long Papers)}, pages 4140--4170, Toronto, Canada. Association for Computational Linguistics.

\bibitem[{Maas et~al.(2011)Maas, Daly, Pham, Huang, Ng, and Potts}]{maas-EtAl:2011:ACL-HLT2011}
Andrew~L. Maas, Raymond~E. Daly, Peter~T. Pham, Dan Huang, Andrew~Y. Ng, and Christopher Potts. 2011.
\newblock \href {http://www.aclweb.org/anthology/P11-1015} {Learning word vectors for sentiment analysis}.
\newblock In \emph{Proceedings of the 49th Annual Meeting of the Association for Computational Linguistics: Human Language Technologies}, pages 142--150, Portland, Oregon, USA. Association for Computational Linguistics.

\bibitem[{Magar and Schwartz(2022)}]{magar-schwartz-2022-data}
Inbal Magar and Roy Schwartz. 2022.
\newblock \href {https://doi.org/10.18653/v1/2022.acl-short.18} {Data contamination: From memorization to exploitation}.
\newblock In \emph{Proceedings of the 60th Annual Meeting of the Association for Computational Linguistics (Volume 2: Short Papers)}, pages 157--165, Dublin, Ireland. Association for Computational Linguistics.

\bibitem[{Marques et~al.(2021)Marques, Matsubara, Nakamura, Ferreira, Wiese, Gadelha, Zaina, Redmiles, and Conte}]{Marques2021-ei}
Leonardo Marques, Patr{\'\i}cia~Gomes Matsubara, Walter~Takashi Nakamura, Bruna~Moraes Ferreira, Igor~Scaliante Wiese, Bruno~Freitas Gadelha, Luciana~Martinez Zaina, David Redmiles, and Tayana~Uch{\^o}a Conte. 2021.
\newblock Understanding {UX} better: A new technique to go beyond emotion assessment.
\newblock \emph{Sensors (Basel)}, 21(21):7183.

\bibitem[{Marzi et~al.(2024)Marzi, Balzano, and Marchiori}]{MARZI2024102545}
Giacomo Marzi, Marco Balzano, and Davide Marchiori. 2024.
\newblock \href {https://doi.org/https://doi.org/10.1016/j.mex.2023.102545} {K-alpha calculator–krippendorff's alpha calculator: A user-friendly tool for computing krippendorff's alpha inter-rater reliability coefficient}.
\newblock \emph{MethodsX}, 12:102545.

\bibitem[{Maxwell et~al.(2015)Maxwell, Lau, and Howard}]{Maxwell2015-bi}
Scott~E Maxwell, Michael~Y Lau, and George~S Howard. 2015.
\newblock Is psychology suffering from a replication crisis? what does ``failure to replicate'' really mean?
\newblock \emph{Am. Psychol.}, 70(6):487--498.

\bibitem[{Maynez et~al.(2020)Maynez, Narayan, Bohnet, and McDonald}]{maynez-etal-2020-faithfulness}
Joshua Maynez, Shashi Narayan, Bernd Bohnet, and Ryan McDonald. 2020.
\newblock \href {https://doi.org/10.18653/v1/2020.acl-main.173} {On faithfulness and factuality in abstractive summarization}.
\newblock In \emph{Proceedings of the 58th Annual Meeting of the Association for Computational Linguistics}, pages 1906--1919, Online. Association for Computational Linguistics.

\bibitem[{McCoy et~al.(2019)McCoy, Pavlick, and Linzen}]{mccoy-etal-2019-right}
Tom McCoy, Ellie Pavlick, and Tal Linzen. 2019.
\newblock \href {https://doi.org/10.18653/v1/P19-1334} {Right for the wrong reasons: Diagnosing syntactic heuristics in natural language inference}.
\newblock In \emph{Proceedings of the 57th Annual Meeting of the Association for Computational Linguistics}, pages 3428--3448, Florence, Italy. Association for Computational Linguistics.

\bibitem[{McHugh(2012)}]{McHugh2012-iy}
Mary~L McHugh. 2012.
\newblock Interrater reliability: the kappa statistic.
\newblock \emph{Biochem. Med. (Zagreb)}, 22(3):276--282.

\bibitem[{Mei et~al.(2023)Mei, Levy, and Wang}]{mei-etal-2023-assert}
Alex Mei, Sharon Levy, and William Wang. 2023.
\newblock \href {https://doi.org/10.18653/v1/2023.findings-emnlp.388} {{ASSERT}: Automated safety scenario red teaming for evaluating the robustness of large language models}.
\newblock In \emph{Findings of the Association for Computational Linguistics: EMNLP 2023}, pages 5831--5847, Singapore. Association for Computational Linguistics.

\bibitem[{Mirnig et~al.(2015)Mirnig, Meschtscherjakov, Wurhofer, Meneweger, and Tscheligi}]{10.1145/2702613.2732511}
Alexander~G. Mirnig, Alexander Meschtscherjakov, Daniela Wurhofer, Thomas Meneweger, and Manfred Tscheligi. 2015.
\newblock \href {https://doi.org/10.1145/2702613.2732511} {A formal analysis of the iso 9241-210 definition of user experience}.
\newblock In \emph{Proceedings of the 33rd Annual ACM Conference Extended Abstracts on Human Factors in Computing Systems}, CHI EA '15, page 437–450, New York, NY, USA. Association for Computing Machinery.

\bibitem[{Narayan et~al.(2018)Narayan, Cohen, and Lapata}]{narayan-etal-2018-dont}
Shashi Narayan, Shay~B. Cohen, and Mirella Lapata. 2018.
\newblock \href {https://doi.org/10.18653/v1/D18-1206} {Don{'}t give me the details, just the summary! topic-aware convolutional neural networks for extreme summarization}.
\newblock In \emph{Proceedings of the 2018 Conference on Empirical Methods in Natural Language Processing}, pages 1797--1807, Brussels, Belgium. Association for Computational Linguistics.

\bibitem[{Nenkova and Passonneau(2004)}]{nenkova-passonneau-2004-evaluating}
Ani Nenkova and Rebecca Passonneau. 2004.
\newblock \href {https://aclanthology.org/N04-1019} {Evaluating content selection in summarization: The pyramid method}.
\newblock In \emph{Proceedings of the Human Language Technology Conference of the North {A}merican Chapter of the Association for Computational Linguistics: {HLT}-{NAACL} 2004}, pages 145--152, Boston, Massachusetts, USA. Association for Computational Linguistics.

\bibitem[{Niranjanamurthy et~al.(2014)Niranjanamurthy, Nagaraj, Gattu, and Shetty}]{niranjanamurthy2014research}
M~Niranjanamurthy, Archikam Nagaraj, Himaja Gattu, and Puneeth~K Shetty. 2014.
\newblock Research study on importance of usability testing/user experience (ux) testing.
\newblock \emph{International Journal of Computer Science and Mobile Computing}, 3(10):78--85.

\bibitem[{Nisbett and Wilson(1977)}]{halo}
Richard Nisbett and Timothy Wilson. 1977.
\newblock \href {https://doi.org/10.1037/0022-3514.35.4.250} {The halo effect: Evidence for unconscious alteration of judgments}.
\newblock \emph{Journal of Personality and Social Psychology}, 35:250--256.

\bibitem[{O'Neill(2017)}]{10.3389/fpsyg.2017.00777}
Thomas~A. O'Neill. 2017.
\newblock \href {https://doi.org/10.3389/fpsyg.2017.00777} {An overview of interrater agreement on likert scales for researchers and practitioners}.
\newblock \emph{Frontiers in Psychology}, 8.

\bibitem[{Perez et~al.(2022)Perez, Huang, Song, Cai, Ring, Aslanides, Glaese, McAleese, and Irving}]{perez-etal-2022-red}
Ethan Perez, Saffron Huang, Francis Song, Trevor Cai, Roman Ring, John Aslanides, Amelia Glaese, Nat McAleese, and Geoffrey Irving. 2022.
\newblock \href {https://doi.org/10.18653/v1/2022.emnlp-main.225} {Red teaming language models with language models}.
\newblock In \emph{Proceedings of the 2022 Conference on Empirical Methods in Natural Language Processing}, pages 3419--3448, Abu Dhabi, United Arab Emirates. Association for Computational Linguistics.

\bibitem[{Pommeranz et~al.(2012)Pommeranz, Broekens, Wiggers, Brinkman, and Jonker}]{Pommeranz2012}
Alina Pommeranz, Joost Broekens, Pascal Wiggers, Willem-Paul Brinkman, and Catholijn~M. Jonker. 2012.
\newblock \href {https://doi.org/10.1007/s11257-011-9116-6} {Designing interfaces for explicit preference elicitation: a user-centered investigation of preference representation and elicitation process}.
\newblock \emph{User Modeling and User-Adapted Interaction}, 22(4):357--397.

\bibitem[{Pradhan et~al.(2022)Pradhan, Schaekermann, and Lease}]{Pradhan2022-wq}
Vivek~Krishna Pradhan, Mike Schaekermann, and Matthew Lease. 2022.
\newblock In search of ambiguity: A three-stage workflow design to clarify annotation guidelines for crowd workers.
\newblock \emph{Front. Artif. Intell.}, 5:828187.

\bibitem[{Pre\ss{}ler et~al.(2023)Pre\ss{}ler, Schmid, and Hurtienne}]{10.1145/3544549.3585739}
Jan Pre\ss{}ler, Lukas Schmid, and J\"{o}rn Hurtienne. 2023.
\newblock \href {https://doi.org/10.1145/3544549.3585739} {Statistically controlling for processing fluency reduces the aesthetic-usability effect}.
\newblock In \emph{Extended Abstracts of the 2023 CHI Conference on Human Factors in Computing Systems}, CHI EA '23, New York, NY, USA. Association for Computing Machinery.

\bibitem[{Radharapu et~al.(2023)Radharapu, Robinson, Aroyo, and Lahoti}]{radharapu-etal-2023-aart}
Bhaktipriya Radharapu, Kevin Robinson, Lora Aroyo, and Preethi Lahoti. 2023.
\newblock \href {https://doi.org/10.18653/v1/2023.emnlp-industry.37} {{AART}: {AI}-assisted red-teaming with diverse data generation for new {LLM}-powered applications}.
\newblock In \emph{Proceedings of the 2023 Conference on Empirical Methods in Natural Language Processing: Industry Track}, pages 380--395, Singapore. Association for Computational Linguistics.

\bibitem[{Reber(2011)}]{10.1093/acprof:oso/9780199732142.003.0055}
Rolf Reber. 2011.
\newblock \href {https://doi.org/10.1093/acprof:oso/9780199732142.003.0055} {{223Processing Fluency, Aesthetic Pleasure, and Culturally Shared Taste}}.
\newblock In \emph{{Aesthetic Science: Connecting Minds, Brains, and Experience}}. Oxford University Press.

\bibitem[{Reber and Schwarz(1999)}]{1999-11900-005}
Rolf Reber and Norbert Schwarz. 1999.
\newblock \href {https://doi.org/10.1006/ccog.1999.0386} {Effects of perceptual fluency on judgments of truth.}
\newblock \emph{Consciousness and Cognition: An International Journal}, 8(3):338--342.

\bibitem[{Ribeiro et~al.(2020)Ribeiro, Wu, Guestrin, and Singh}]{ribeiro-etal-2020-beyond}
Marco~Tulio Ribeiro, Tongshuang Wu, Carlos Guestrin, and Sameer Singh. 2020.
\newblock \href {https://doi.org/10.18653/v1/2020.acl-main.442} {Beyond accuracy: Behavioral testing of {NLP} models with {C}heck{L}ist}.
\newblock In \emph{Proceedings of the 58th Annual Meeting of the Association for Computational Linguistics}, pages 4902--4912, Online. Association for Computational Linguistics.

\bibitem[{Rogers(2021)}]{rogers-2021-changing}
Anna Rogers. 2021.
\newblock \href {https://doi.org/10.18653/v1/2021.acl-long.170} {Changing the world by changing the data}.
\newblock In \emph{Proceedings of the 59th Annual Meeting of the Association for Computational Linguistics and the 11th International Joint Conference on Natural Language Processing (Volume 1: Long Papers)}, pages 2182--2194, Online. Association for Computational Linguistics.

\bibitem[{Rottger et~al.(2022)Rottger, Vidgen, Hovy, and Pierrehumbert}]{rottger-etal-2022-two}
Paul Rottger, Bertie Vidgen, Dirk Hovy, and Janet Pierrehumbert. 2022.
\newblock \href {https://doi.org/10.18653/v1/2022.naacl-main.13} {Two contrasting data annotation paradigms for subjective {NLP} tasks}.
\newblock In \emph{Proceedings of the 2022 Conference of the North American Chapter of the Association for Computational Linguistics: Human Language Technologies}, pages 175--190, Seattle, United States. Association for Computational Linguistics.

\bibitem[{Rusert et~al.(2022)Rusert, Shafiq, and Srinivasan}]{rusert-etal-2022-robustness}
Jonathan Rusert, Zubair Shafiq, and Padmini Srinivasan. 2022.
\newblock \href {https://doi.org/10.18653/v1/2022.acl-long.513} {On the robustness of offensive language classifiers}.
\newblock In \emph{Proceedings of the 60th Annual Meeting of the Association for Computational Linguistics (Volume 1: Long Papers)}, pages 7424--7438, Dublin, Ireland. Association for Computational Linguistics.

\bibitem[{Sainz et~al.(2023)Sainz, Campos, Garc{\'\i}a-Ferrero, Etxaniz, de~Lacalle, and Agirre}]{sainz-etal-2023-nlp}
Oscar Sainz, Jon Campos, Iker Garc{\'\i}a-Ferrero, Julen Etxaniz, Oier~Lopez de~Lacalle, and Eneko Agirre. 2023.
\newblock \href {https://doi.org/10.18653/v1/2023.findings-emnlp.722} {{NLP} evaluation in trouble: On the need to measure {LLM} data contamination for each benchmark}.
\newblock In \emph{Findings of the Association for Computational Linguistics: EMNLP 2023}, pages 10776--10787, Singapore. Association for Computational Linguistics.

\bibitem[{Santhanam et~al.(2020)Santhanam, Karduni, and Shaikh}]{10.1145/3313831.3376318}
Sashank Santhanam, Alireza Karduni, and Samira Shaikh. 2020.
\newblock \href {https://doi.org/10.1145/3313831.3376318} {Studying the effects of cognitive biases in evaluation of conversational agents}.
\newblock In \emph{Proceedings of the 2020 CHI Conference on Human Factors in Computing Systems}, CHI '20, page 1–13, New York, NY, USA. Association for Computing Machinery.

\bibitem[{Schoch et~al.(2020)Schoch, Yang, and Ji}]{schoch-etal-2020-problem}
Stephanie Schoch, Diyi Yang, and Yangfeng Ji. 2020.
\newblock \href {https://aclanthology.org/2020.evalnlgeval-1.2} {{``}this is a problem, don{'}t you agree?{''} framing and bias in human evaluation for natural language generation}.
\newblock In \emph{Proceedings of the 1st Workshop on Evaluating NLG Evaluation}, pages 10--16, Online (Dublin, Ireland). Association for Computational Linguistics.

\bibitem[{Schwarz(2006)}]{10.1162/daed.2006.135.2.136}
Norbert Schwarz. 2006.
\newblock \href {https://doi.org/10.1162/daed.2006.135.2.136} {{on judgments of truth \& beauty}}.
\newblock \emph{Daedalus}, 135(2):136--138.

\bibitem[{Sclar et~al.(2024)Sclar, Choi, Tsvetkov, and Suhr}]{sclar2024quantifying}
Melanie Sclar, Yejin Choi, Yulia Tsvetkov, and Alane Suhr. 2024.
\newblock \href {https://openreview.net/forum?id=RIu5lyNXjT} {Quantifying language models' sensitivity to spurious features in prompt design or: How i learned to start worrying about prompt formatting}.
\newblock In \emph{The Twelfth International Conference on Learning Representations}.

\bibitem[{Sonderegger and Sauer(2010)}]{SONDEREGGER2010403}
Andreas Sonderegger and Juergen Sauer. 2010.
\newblock \href {https://doi.org/https://doi.org/10.1016/j.apergo.2009.09.002} {The influence of design aesthetics in usability testing: Effects on user performance and perceived usability}.
\newblock \emph{Applied Ergonomics}, 41(3):403--410.
\newblock Special Section: Recycling centres and waste handling – a workplace for employees and users.

\bibitem[{Sukumar and Metoyer(2018)}]{8634261_Unbiased_Replication_Studies}
Poorna~Talkad Sukumar and Ronald Metoyer. 2018.
\newblock \href {https://doi.org/10.1109/BELIV.2018.8634261} {Towards designing unbiased replication studies in information visualization}.
\newblock In \emph{2018 IEEE Evaluation and Beyond - Methodological Approaches for Visualization (BELIV)}, pages 93--101.

\bibitem[{Sun et~al.(2024)Sun, Huang, Wang, Wu, Zhang, Gao, Huang, Lyu, Zhang, Li, Liu, Liu, Wang, Zhang, Kailkhura, Xiong, Xiao, Li, Xing, Huang, Liu, Ji, Wang, Zhang, Yao, Kellis, Zitnik, Jiang, Bansal, Zou, Pei, Liu, Gao, Han, Zhao, Tang, Wang, Mitchell, Shu, Xu, Chang, He, Huang, Backes, Gong, Yu, Chen, Gu, Xu, Ying, Ji, Jana, Chen, Liu, Zhou, Wang, Li, Zhang, Wang, Xie, Chen, Wang, Liu, Ye, Cao, Chen, and Zhao}]{sun2024trustllm}
Lichao Sun, Yue Huang, Haoran Wang, Siyuan Wu, Qihui Zhang, Chujie Gao, Yixin Huang, Wenhan Lyu, Yixuan Zhang, Xiner Li, Zhengliang Liu, Yixin Liu, Yijue Wang, Zhikun Zhang, Bhavya Kailkhura, Caiming Xiong, Chaowei Xiao, Chunyuan Li, Eric Xing, Furong Huang, Hao Liu, Heng Ji, Hongyi Wang, Huan Zhang, Huaxiu Yao, Manolis Kellis, Marinka Zitnik, Meng Jiang, Mohit Bansal, James Zou, Jian Pei, Jian Liu, Jianfeng Gao, Jiawei Han, Jieyu Zhao, Jiliang Tang, Jindong Wang, John Mitchell, Kai Shu, Kaidi Xu, Kai-Wei Chang, Lifang He, Lifu Huang, Michael Backes, Neil~Zhenqiang Gong, Philip~S. Yu, Pin-Yu Chen, Quanquan Gu, Ran Xu, Rex Ying, Shuiwang Ji, Suman Jana, Tianlong Chen, Tianming Liu, Tianyi Zhou, William Wang, Xiang Li, Xiangliang Zhang, Xiao Wang, Xing Xie, Xun Chen, Xuyu Wang, Yan Liu, Yanfang Ye, Yinzhi Cao, Yong Chen, and Yue Zhao. 2024.
\newblock \href {http://arxiv.org/abs/2401.05561} {Trustllm: Trustworthiness in large language models}.

\bibitem[{Taherdoost(2019)}]{RePEc:hal:journl:hal-02557308}
Hamed Taherdoost. 2019.
\newblock \href {https://ideas.repec.org/p/hal/journl/hal-02557308.html} {{What Is the Best Response Scale for Survey and Questionnaire Design; Review of Different Lengths of Rating Scale / Attitude Scale / Likert Scale}}.
\newblock Post-Print hal-02557308, HAL.

\bibitem[{Tedeschi et~al.(2023)Tedeschi, Bos, Declerck, Haji{\v{c}}, Hershcovich, Hovy, Koller, Krek, Schockaert, Sennrich, Shutova, and Navigli}]{tedeschi-etal-2023-whats}
Simone Tedeschi, Johan Bos, Thierry Declerck, Jan Haji{\v{c}}, Daniel Hershcovich, Eduard Hovy, Alexander Koller, Simon Krek, Steven Schockaert, Rico Sennrich, Ekaterina Shutova, and Roberto Navigli. 2023.
\newblock \href {https://doi.org/10.18653/v1/2023.acl-long.697} {What{'}s the meaning of superhuman performance in today{'}s {NLU}?}
\newblock In \emph{Proceedings of the 61st Annual Meeting of the Association for Computational Linguistics (Volume 1: Long Papers)}, pages 12471--12491, Toronto, Canada. Association for Computational Linguistics.

\bibitem[{ten Hove et~al.(2018)ten Hove, Jorgensen, and van~der Ark}]{10.1007/978-3-319-77249-3_6}
Debby ten Hove, Terrence~D. Jorgensen, and L.~Andries van~der Ark. 2018.
\newblock On the usefulness of interrater reliability coefficients.
\newblock In \emph{Quantitative Psychology}, pages 67--75, Cham. Springer International Publishing.

\bibitem[{Tsai and Thomas(2011)}]{tsai2011does}
Claire~I Tsai and Manoj Thomas. 2011.
\newblock When does feeling of fluency matter? how abstract and concrete thinking influence fluency effects.
\newblock \emph{Psychological Science}, 22(3):348--354.

\bibitem[{Tuch et~al.(2012)Tuch, Roth, Hornbæk, Opwis, and Bargas-Avila}]{TUCH20121596}
Alexandre~N. Tuch, Sandra~P. Roth, Kasper Hornbæk, Klaus Opwis, and Javier~A. Bargas-Avila. 2012.
\newblock \href {https://doi.org/https://doi.org/10.1016/j.chb.2012.03.024} {Is beautiful really usable? toward understanding the relation between usability, aesthetics, and affect in hci}.
\newblock \emph{Computers in Human Behavior}, 28(5):1596--1607.

\bibitem[{Tullis and Albert(2013)}]{TULLIS201341}
Tom Tullis and Bill Albert. 2013.
\newblock \href {https://doi.org/https://doi.org/10.1016/B978-0-12-415781-1.00003-0} {Chapter 3 - planning}.
\newblock In Tom Tullis and Bill Albert, editors, \emph{Measuring the User Experience (Second Edition)}, second edition edition, Interactive Technologies, pages 41--62. Morgan Kaufmann, Boston.

\bibitem[{Tversky and Kahneman(1974)}]{doi:10.1126/science.185.4157.1124}
Amos Tversky and Daniel Kahneman. 1974.
\newblock \href {https://doi.org/10.1126/science.185.4157.1124} {Judgment under uncertainty: Heuristics and biases}.
\newblock \emph{Science}, 185(4157):1124--1131.

\bibitem[{Vakili and Dalianis(2023)}]{vakili-dalianis-2023-using}
Thomas Vakili and Hercules Dalianis. 2023.
\newblock \href {https://aclanthology.org/2023.nodalida-1.33} {Using membership inference attacks to evaluate privacy-preserving language modeling fails for pseudonymizing data}.
\newblock In \emph{Proceedings of the 24th Nordic Conference on Computational Linguistics (NoDaLiDa)}, pages 318--323, T{\'o}rshavn, Faroe Islands. University of Tartu Library.

\bibitem[{{van der Lee} et~al.(2021){van der Lee}, Gatt, {van Miltenburg}, and Krahmer}]{VANDERLEE2021101151}
Chris {van der Lee}, Albert Gatt, Emiel {van Miltenburg}, and Emiel Krahmer. 2021.
\newblock \href {https://doi.org/https://doi.org/10.1016/j.csl.2020.101151} {Human evaluation of automatically generated text: Current trends and best practice guidelines}.
\newblock \emph{Computer Speech \& Language}, 67:101151.

\bibitem[{van~der Lee et~al.(2019)van~der Lee, Gatt, van Miltenburg, Wubben, and Krahmer}]{van-der-lee-etal-2019-best}
Chris van~der Lee, Albert Gatt, Emiel van Miltenburg, Sander Wubben, and Emiel Krahmer. 2019.
\newblock \href {https://doi.org/10.18653/v1/W19-8643} {Best practices for the human evaluation of automatically generated text}.
\newblock In \emph{Proceedings of the 12th International Conference on Natural Language Generation}, pages 355--368, Tokyo, Japan. Association for Computational Linguistics.

\bibitem[{vocabulary.com()}]{vocabulary}
vocabulary.com.
\newblock \url{https://www.vocabulary.com/dictionary/differentiate}.
\newblock Accessed: 2024-4-5.

\bibitem[{Wang et~al.(2019{\natexlab{a}})Wang, Pruksachatkun, Nangia, Singh, Michael, Hill, Levy, and Bowman}]{wang2019superglue}
Alex Wang, Yada Pruksachatkun, Nikita Nangia, Amanpreet Singh, Julian Michael, Felix Hill, Omer Levy, and Samuel Bowman. 2019{\natexlab{a}}.
\newblock \href {https://proceedings.neurips.cc/paper_files/paper/2019/file/4496bf24afe7fab6f046bf4923da8de6-Paper.pdf} {Superglue: A stickier benchmark for general-purpose language understanding systems}.
\newblock In \emph{Advances in Neural Information Processing Systems}, volume~32. Curran Associates, Inc.

\bibitem[{Wang et~al.(2019{\natexlab{b}})Wang, Singh, Michael, Hill, Levy, and Bowman}]{wang2019glue}
Alex Wang, Amanpreet Singh, Julian Michael, Felix Hill, Omer Levy, and Samuel~R. Bowman. 2019{\natexlab{b}}.
\newblock \href {https://openreview.net/forum?id=rJ4km2R5t7} {{GLUE:} {A} multi-task benchmark and analysis platform for natural language understanding}.
\newblock In \emph{7th International Conference on Learning Representations, {ICLR} 2019, New Orleans, LA, USA, May 6-9, 2019}. OpenReview.net.

\bibitem[{White(2013)}]{white2013beliefs}
Ryen~W. White. 2013.
\newblock \href {https://www.microsoft.com/en-us/research/publication/beliefs-biases-web-search/} {Beliefs and biases in web search}.
\newblock In \emph{36th Annual International ACM SIGIR Conference on Research and Development in Information Retrieval (SIGIR 2013), Dublin, Ireland}, pages 3--12.
\newblock Best Paper Award.

\bibitem[{Willits et~al.(2016)Willits, Theodori, and Luloff}]{willits2016another}
Fern~K Willits, Gene~L Theodori, and AE~Luloff. 2016.
\newblock Another look at likert scales.
\newblock \emph{Journal of Rural Social Sciences}, 31(3):6.

\bibitem[{Wu and Quinn(2017)}]{Wu_Quinn_2017}
Meng-Han Wu and Alexander Quinn. 2017.
\newblock \href {https://doi.org/10.1609/hcomp.v5i1.13317} {Confusing the crowd: Task instruction quality on amazon mechanical turk}.
\newblock \emph{Proceedings of the AAAI Conference on Human Computation and Crowdsourcing}, 5(1):206--215.

\end{thebibliography}

\onecolumn
\appendix
\section {Appendix}

\subsection{ConSiDERS-The-Human Evaluation Checklist}\label{sec:checklist}
\subsubsection*{Consistency}
{\small
\noindent\textit{Ill-defined or complex evaluation guidelines}
\begin{enumerate}[noitemsep,topsep=1pt]
    \item Were expert annotators independently able to follow the annotation guidelines and achieve higher inter-annotator agreement? 
    \item Is there a mechanism for evaluators to report issues such as ambiguity in the guideline? 
 
\end{enumerate}

\noindent\textit{High task complexity}
\begin{enumerate}[noitemsep,topsep=1pt]
    \setcounter{enumi}{2}
    \item Can you simplify the task by breaking down the task into easier tasks?
    \item Are you asking multiple questions in a single task? If yes, split each question into separate tasks so that \textbf{a)} the answer to one question doesn't bias another one, \textbf{b)} the annotators feel positively motivated that they can complete a given task fast.
\end{enumerate}

\noindent\textit{Ill-suited evaluators}
\begin{enumerate}[noitemsep,topsep=1pt]
 \setcounter{enumi}{4}
    \item Is there a qualification exam for evaluators?
    \item Do you insert exam and attention check quality control metrics during evaluation to identify the quality of individual annotators as part of each batch of evaluation?
    \item Does your task need fresh annotators? For instance, if your task is to measure how good the LLM's instructions are, you need to make sure that the annotators are new and not used to how to complete the task.
\end{enumerate}

\noindent\textit{Small number of evaluators and/or test set}
\begin{enumerate}[noitemsep,topsep=1pt]
    \setcounter{enumi}{7}
    \item Is it possible to include more evaluators?
    \item  Is it possible to increase the size of the test set?
\end{enumerate}

\noindent\textit{Rating scales such as the Likert}
\begin{enumerate}[noitemsep,topsep=1pt]
\setcounter{enumi}{9}
    \item Are you trying to measure perception or non-perception-based aspects such as truthfulness? If you are measuring non-perception-based metrics,  avoid the Likert scoring, and modify the task to collect many more objective metrics such as extracting facts and verifying each fact.
\end{enumerate}

\noindent\textit{Inter-rater agreement}
\begin{enumerate}[noitemsep,topsep=1pt]
\setcounter{enumi}{10}
 \item Which primary IRA measure did you use?
 \item Does the primary measure take into account how the evaluation is designed, such as the aspects defined by \citet{GISEV2013330}?
 \item Do you expect your task to be inherently unbalanced? If not, do you report baseline percentage agreement to verify if the observed chance estimation lowers the overall IRA?

\item Is the item-wise IRA higher for some items and not the others? The ones with lower IRA might be pointing to higher complexity tasks. Report on the distribution of items IRA to troubleshoot the problem.

\item Are the qualifications of the human evaluators similar? If not, the disparity in the evaluators' skills can lead to lower IRA.

\item Do you continuously measure IRA for each evaluation? If yes, did you observe a sudden change in IRA? It might be due to changes in annotators (e.g., adding new annotators) or changes in guidelines or tasks. New guidelines and tasks take a few iterations to settle.
  
\end{enumerate}

\subsubsection*{Scoring Criteria}

\begin{enumerate}[noitemsep,topsep=1pt]
\item Is the model evaluated on typical dimensions: Fluency, Coherence, Relevance, Factuality?
\item What multi-dimensional domain-specific criteria were the model evaluated on?
\item What are the responsible AI criteria the model was evaluated on?
\end{enumerate}

\subsubsection*{Differentiating}

\begin{enumerate}
   [noitemsep,topsep=1pt]
\item How many test cases (test examples) did you use to evaluate each of the criteria?
\item What were the end user use cases, e.g., legal document summarization, were tested? If so, report the end user test cases, their corresponding number of tests, and the scores per user case per criteria.
\item Do you suspect that some of the test cases may have already been used during LLM training? If not sure, answer not sure.  If yes, where possible, report the percentage of test cases that may have been impacted.
\item Did you run robustness tests to evaluate model weaknesses? What were the scenarios covered, e.g.,  semantic preserving perturbations - such as sensitivity to white spaces, synonyms, etc?
\end{enumerate}

\subsubsection*{User Experience}

\begin{enumerate}[noitemsep,topsep=1pt]
\item Were rating denoising algorithms applied on the rating-based metrics, such as the Likert, to account for the cognitive uncertainty of the human evaluators?
\item Did you split the content into atomic facts for criteria, such as factual completeness and truthfulness,  that require inspecting individual traits of the model-generated text? If so, detail the criteria. 
\item Is the Likert scale used appropriately to measure perception-based aspects such as readability?
\item During human evaluation, were examples shuffled properly so that the evaluator cannot tell which model generated which example?
\item Did you test the actual usability of the model’s output, in the context of the end-user application (aka extrinsic evaluation)? If so, briefly describe the details.
\end{enumerate}

\subsubsection*{Responsible}

\begin{enumerate}[noitemsep,topsep=1pt]
\item Was safety testing performed? If yes, how many test cases were used and what were the test scenarios? 
\item Was privacy testing performed? If yes, how many test cases were used and what were the test scenarios?
\item Was the model tested for bias? If yes, what subgroups such as gender or disability was the model evaluated for?
\item How many evaluators were used? What was the diversity of evaluators or several evaluators grouped by aspects such as education qualifications, race, religion, gender, age groups, and nationalities?
\end{enumerate}

\subsubsection*{Scalability}

\begin{enumerate}[noitemsep,topsep=1pt]
\item Were parts of the human evaluation automated? If yes, which aspects did you attempt to automate, and what types of automated metrics were used? 
\item If you used LLM-based automation, please provide the details of the LLM and prompts. 
\item Did you optimize the usability aspects for the annotators to reduce annotation time? If so, provide a summary of how this was achieved.
\end{enumerate}

}

\subsection{Meta-analysis of papers available in ACL Anthology}
\label{sec:metaanalysis}

\subsubsection{Papers with  ``human'' and  ``eval'' in either abstract or title}

\begin{figure}[h!]
    \centering
    \includegraphics[width=1\linewidth]{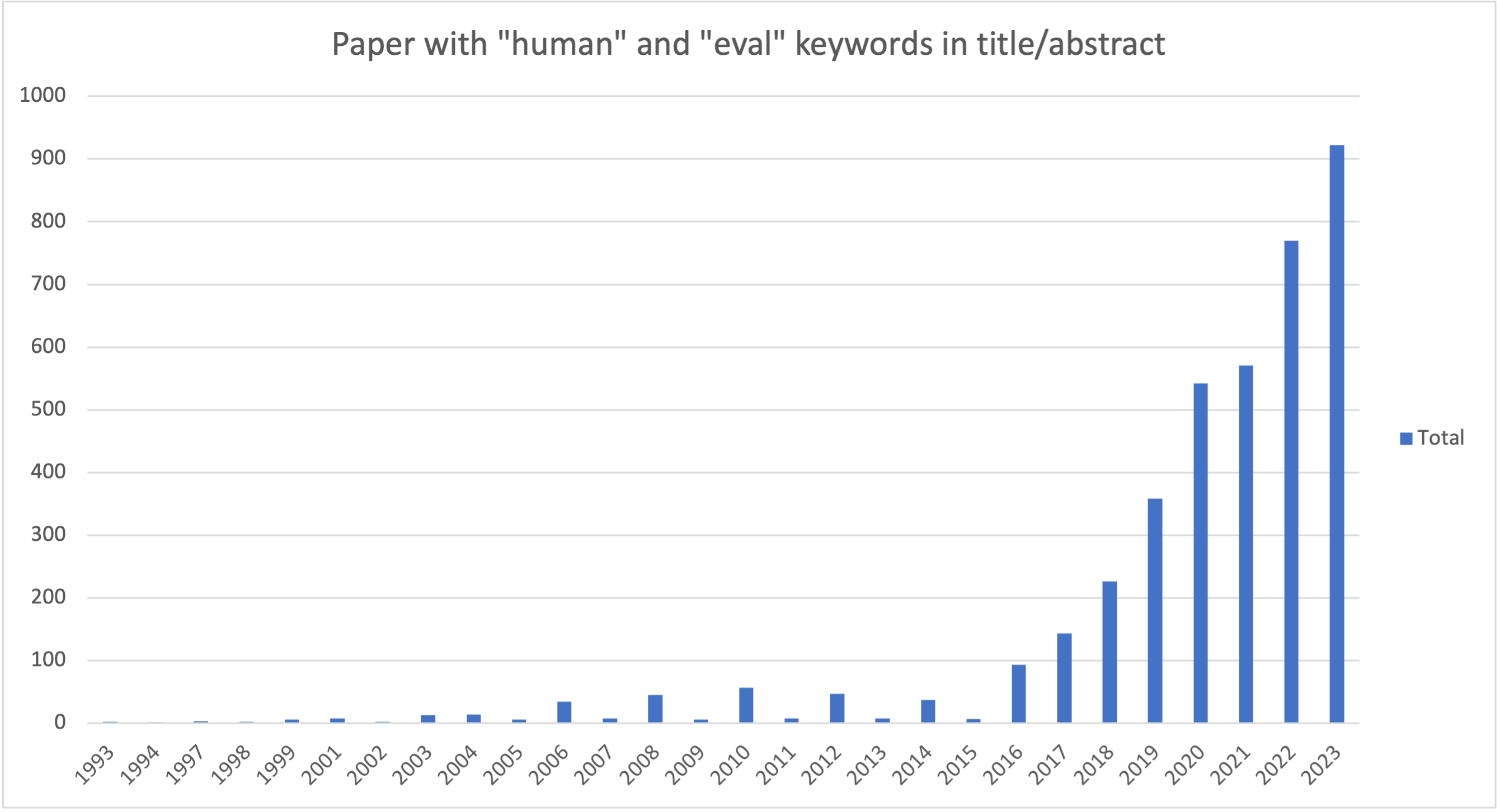}
    \caption{Yearly publications of papers with keywords  ``human'' and  ``eval'' in title/abstract}
    \label{app:fig:humanevalkeywordstitleabstract}
\end{figure}

We search for abstracts or titles containing keywords ``human'' and  ``eval'', with $\approx$3900 papers, where 900 of those published in 2023 as detailed in Figure~\ref{app:fig:humanevalkeywordstitleabstract}.

\begin{lstlisting}
       (title|abstract=human and title|abstract=eval)    
\end{lstlisting}

\subsubsection{Papers with  ``human'' and  ``eval'' in the title only}

\begin{figure}[h]
    \centering
    \includegraphics[width=1\linewidth]{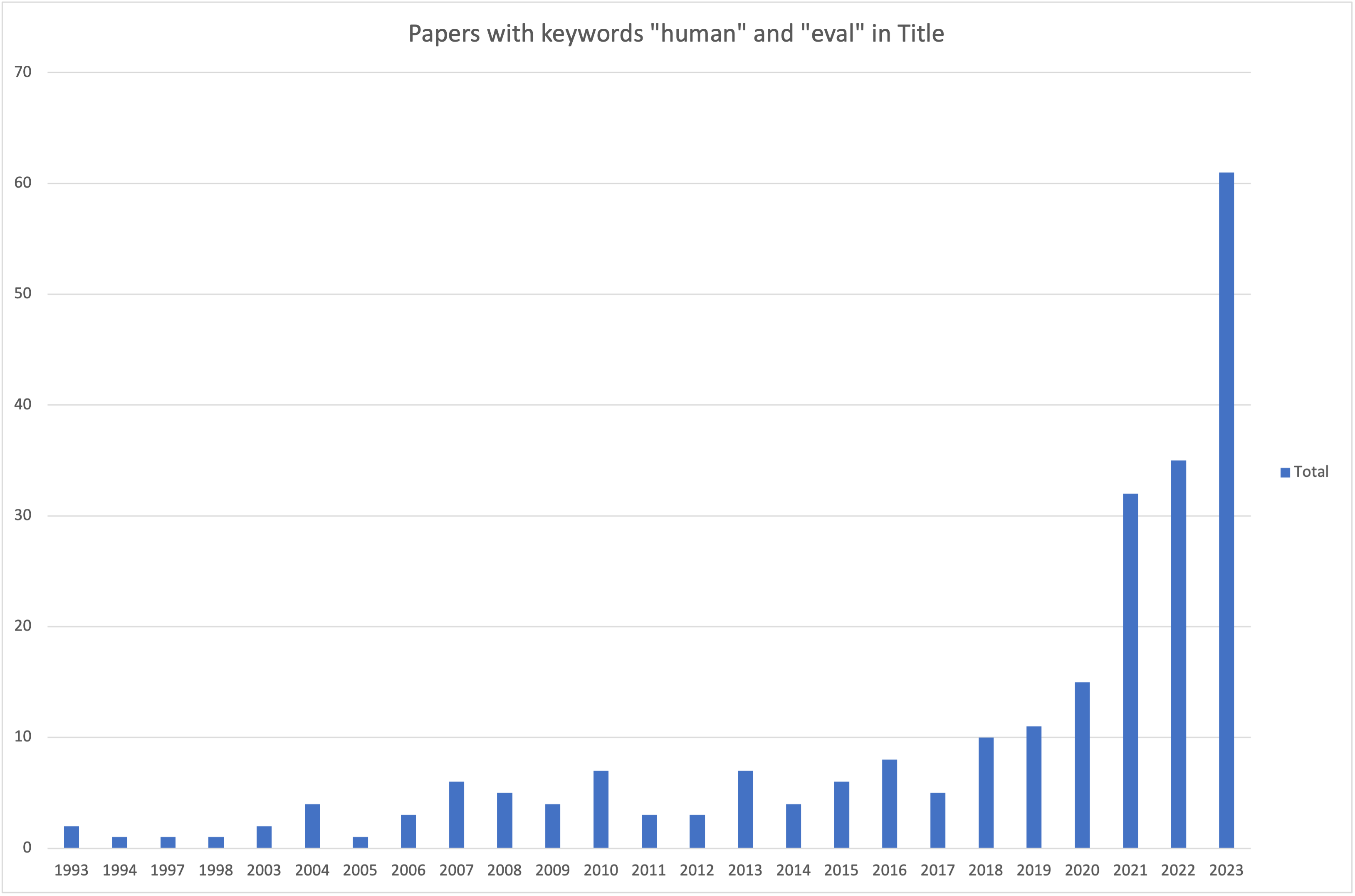}
    \caption{Yearly publications of papers with keywords  ``human'' and  ``eval'' in title}
    \label{app:fig:humanevalkeywordstitle}
\end{figure}

We search for titles containing keywords ``human'' and  ``eval'', this results in around 238 papers as shown in Figure~\ref{app:fig:humanevalkeywordstitle}, reducing the number from ~3900 when we include abstracts as seen above.

\begin{lstlisting}
    (    (title=human and title=eval)     ) 
\end{lstlisting}

\subsubsection{Papers with phrase  ``human  eval'' in either abstract or title}

We search for abstracts or titles containing phrases human evaluation, and human - evaluation. Around 1300 papers were published.

\begin{lstlisting}
    (    (title|abstract=[h|H]uman\s?-?\s?[e|E]val)     ) 
\end{lstlisting}

\subsubsection{Papers with keywords  ``human'' and  ``eval'' in title/abstract and usability keywords in either abstract or title}

We search for titles or abstracts that mention human evaluation and contain usability keywords (usability, hci, user experience, and human computer). This results in 172 papers. The problem with this query is that it is too noisy as a result of looking for ``human'' and  ``eval'' in either the title or abstract. For instance, while results contain some human evaluation, the paper's primary focus is not on the design of human evaluation itself.

\begin{lstlisting}[caption={Jabref search query for human eval and usablity}]
(
    (title|abstract=human and title|abstract=eval)  
    and
    (      (title|abstract=usability) 
        or (title|abstract=[uU]ser\s[eE]xperience) 
        or (title|abstract=user and title|abstract=studies) 
        or (title|abstract=hci) 
        or (title|abstract=human and title|abstract=computer)  
    ) 
) 
\end{lstlisting}

\subsubsection{Papers with phrase  ``human  eval'' in either abstract/title and usability keywords}

\begin{lstlisting}[caption={Jabref search query for phrase human eval in title and usablity in title or abstract}]
(
    (    (title|abstract=[h|H]uman\s?-?\s?[e|E]val)    
    
    ) 
    and
    (      (title|abstract=usability) 
        or (title|abstract=[uU]ser\s[eE]xperience) 
        or (title|abstract=user and title|abstract=studies) 
        or (title|abstract=hci) 
        or (title|abstract=human and title|abstract=computer)  
    ) 
) 
\end{lstlisting}

\subsubsection{Papers with keywords  ``human'' and  ``eval'' in title and usability keywords}\label{app:sec:humanevalkeywordintitlewithusability}

Out of 238 paper keywords human and eval in the title ``(title=human and  title=eval )", only 16 mention usability-related keywords in the title or abstract ACL.

\begin{lstlisting}[caption={Jabref search query for phrase human  in title and usablity in title or abstract}]
(
    (    (title=human and  title=eval   ) )
    and
    (      (title|abstract=usability) 
        or (title|abstract=user and title|abstract=research ) 
        or (title|abstract=user and title|abstract=experience ) 
        or (title|abstract=user and title|abstract=studies) 
        or (title|abstract=hci) 
        or (title|abstract=human and title|abstract=computer)  
    ) 
) 
\end{lstlisting}

\begin{figure}
    \centering
    \includegraphics[width=1\linewidth]{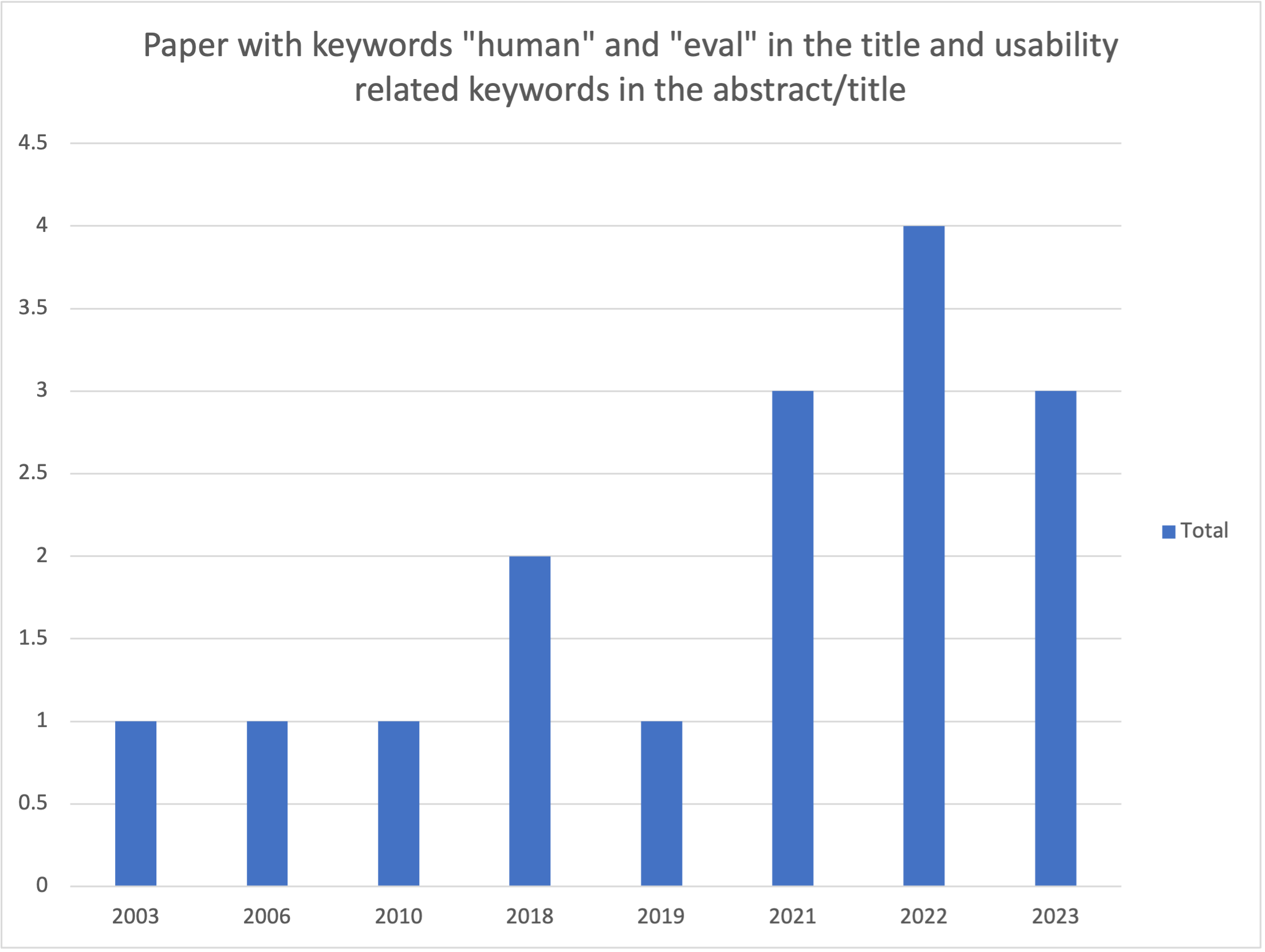}
    \caption{Yearly publications of papers with keywords  ``human'' and  ``eval'' in title and usability keywords in the abstract/title}
    \label{app:fig:humanevaltitle}
\end{figure}

\subsubsection{Paper that attempt to scale or optimise human evaluation}\label{app:sec:humanevalscalecost}

\begin{lstlisting}[caption={Jabref search query for phrase human  in title and cost / scale in title or abstract}]
(
    (    (title=human and  title=eval   ) )
    and
    (      (title|abstract=scal) 
        or (title|abstract=cost ) 
        or (title|abstract=time ) 
    ) 
) 
\end{lstlisting}

\begin{lstlisting}[caption={Jabref search query for auto eval human  in title}]
(
        (title=auto and  title=eval   ) 
) 
\end{lstlisting}

\subsubsection{Paper that attempt to mention responsible AI related terms in their title}\label{app:sec:responsibleAI}

\begin{lstlisting}[caption={Jabref search RAI terms in  title}]
(
    (
        (title=responsib) 
        or (title=fair) 
        or (title=truth)
        or (title=trust)
        or (title=privacy)
        or (title=bias)
        or (title=safe)
    )
    and
    ( not (title=inductive and title=bias))
) 

\end{lstlisting}

\begin{lstlisting}[caption={Jabref search Human eval in title and RAI in asbtract}]
(title=human and  title=eval   ) 
and
(
      (
        (abstract|title=responsib) 
        or (abstract|title=fair) 
        or (abstract|title=truth)
        or (title=trust)
        or (abstract|title=privacy)
        or (abstract|title=bias)
        or (abstract|title=safe)
    )
    and
    ( not (abstract|title=inductive and abstract|title=bias))
) 

\end{lstlisting}

\subsection{Krippendorff's Toy example}\label{app:sec:toyexample}

In this example, we demonstrate with 6 items, 6 raters where each item is rated exactly by 3 raters how simply changing 1 label (row 4, col 2) the Krippendorff's-$\alpha$  drops from 0.7 to 0.24, while percentage agreement only drops from 94.4 to 88.9. In scenario 3, we again just change 1 label, where the \% agreement remains the same at 88\%, but Krippendorff's-$\alpha$ drops from 0.24 to -0.06.

\begin{python}
import numpy as np
import krippendorff
from collections import Counter

def compute_agreement(str_reliability_data):
    reliability_data = []

    for coder in str_reliability_data:
        reliability_data.append([np.nan if l == "*" else int(l)
                                 for l in coder.split()])

    compute_krippendorff(reliability_data)
    compute_percentage_agreement(reliability_data)

def compute_krippendorff(reliability_data):
    agreement = krippendorff.alpha(reliability_data=reliability_data,
                                   level_of_measurement="nominal")
    print(np.matrix(reliability_data))
    print("Krippendorff's alpha for nominal metric: ", round(agreement, 2))

def compute_percentage_agreement(reliability_data):
    """
        Percentage agreement, only for binary values.
    """
    item_wise_data = np.array(reliability_data).T
    item_wise_agreement = []
    print("**Percentage agreement**")
    for item in item_wise_data:
        row_labels = [l for l in item if not np.isnan(l)]
        highest_frequency = Counter(row_labels).most_common(1)[0][1]
        item_percentage = round(100 * highest_frequency / len(row_labels), 2)
        item_wise_agreement.append(item_percentage)

        print(row_labels, item_percentage)

    print("Percentage agreement :", np.mean(item_wise_agreement))

print("***Scenario 1: Baseline ****")
compute_agreement(str_reliability_data=(
    "1    0    *  *  *  *",  # coder A
    "1    *    1  1  1  *",  # coder B
    "1    0    1  1  1  *",  # coder C
    "*    0    1  1  *  1",  # coder D
    "*    *    *  *  1  1",  # coder E
    "*    *    *  *  *  0",  # coder F
))

print()
print("***Scenario 2: Change just 1 label, results in much lower alpha  ****")
compute_agreement(str_reliability_data=(
    "1    0    *  *  *  *",  # coder A
    "1    *    1  1  1  *",  # coder B
    "1    0    1  1  1  *",  # coder C
    "*    1    1  1  *  1",  # coder D
    "*    *    *  *  1  1",  # coder E
    "*    *    *  *  *  0",  # coder F
))

print()
print("***Scenario 3: Same 
compute_agreement(str_reliability_data=(
    "1    1    *  *  *  *",  # coder A
    "1    *    1  1  1  *",  # coder B
    "1    0    1  1  1  *",  # coder C
    "*    1    1  1  *  1",  # coder D
    "*    *    *  *  1  1",  # coder E
    "*    *    *  *  *  0",  # coder F
))

\end{python}
This produces the following output

\begin{lstlisting}
***Scenario 1: Baseline ****
[[ 1.  0. nan nan nan nan]
 [ 1. nan  1.  1.  1. nan]
 [ 1.  0.  1.  1.  1. nan]
 [nan  0.  1.  1. nan  1.]
 [nan nan nan nan  1.  1.]
 [nan nan nan nan nan  0.]]
Krippendorff's alpha for nominal metric:  0.7
**Percentage agreement**
[1.0, 1.0, 1.0] 100.0
[0.0, 0.0, 0.0] 100.0
[1.0, 1.0, 1.0] 100.0
[1.0, 1.0, 1.0] 100.0
[1.0, 1.0, 1.0] 100.0
[1.0, 1.0, 0.0] 66.67
Percentage agreement : 94.445

***Scenario 2: Change just 1 label, results in much lower alpha  ****
[[ 1.  0. nan nan nan nan]
 [ 1. nan  1.  1.  1. nan]
 [ 1.  0.  1.  1.  1. nan]
 [nan  1.  1.  1. nan  1.]
 [nan nan nan nan  1.  1.]
 [nan nan nan nan nan  0.]]
Krippendorff's alpha for nominal metric:  0.24
**Percentage agreement**
[1.0, 1.0, 1.0] 100.0
[0.0, 0.0, 1.0] 66.67
[1.0, 1.0, 1.0] 100.0
[1.0, 1.0, 1.0] 100.0
[1.0, 1.0, 1.0] 100.0
[1.0, 1.0, 0.0] 66.67
Percentage agreement : 88.89

***Scenario 3: Same % agreement as Scenario 2, but much lower alpha  ****
[[ 1.  1. nan nan nan nan]
 [ 1. nan  1.  1.  1. nan]
 [ 1.  0.  1.  1.  1. nan]
 [nan  1.  1.  1. nan  1.]
 [nan nan nan nan  1.  1.]
 [nan nan nan nan nan  0.]]
Krippendorff's alpha for nominal metric:  -0.06
**Percentage agreement**
[1.0, 1.0, 1.0] 100.0
[1.0, 0.0, 1.0] 66.67
[1.0, 1.0, 1.0] 100.0
[1.0, 1.0, 1.0] 100.0
[1.0, 1.0, 1.0] 100.0
[1.0, 1.0, 0.0] 66.67
Percentage agreement : 88.89

\end{lstlisting}

\subsection{Papers in  medical journals that use the Likert scale for evaluating ChatGPT}\label{app:sec:chatgpt}
We searched    medical journals as follows and manually selected 19 (not cherry-picking) papers that used human evaluation the Likert scale to assess ChatGPT. We find that the Likert scale was used to evaluate factual completeness, saliency correctness in 9/19 papers,  4/19 of papers use the Likert scale appropriately to measure user perception and for the rest of 6/19 we were not sure of what the criteria meant, for details see Table~\ref{app:tab:GPTlikert}.

\begin{itemize}
    \item \textbf{Nature \& sub-journals:} Performed search within the nature website using keywords $\langle$gpt medical .
    likert$\rangle$
    \item \textbf{Lancet:} Performed search within the Lancet website  using keywords $\langle$gpt  
    likert$\rangle$.
    \item \textbf{JMIR:} Searched for $\langle$gpt likert, jmir$\rangle$ on Google Scholar and used the first 2 pages of results to filter relavant context.
\end{itemize}

\begin{table*}[htb]
\centering\scriptsize
\renewcommand{\arraystretch}{2}
\begin{tabular}{p{3cm}p{7cm}p{2.5cm}p{1.5cm}}
\toprule
\textbf{Title}                                                                                                                             & \textbf{Relevant quote from paper}                                                                                                                                                                                                                                                                                                                                                                                                                                                                                                                                                                                                                                                                                           & \textbf{Dimensions}                                                                                                                                                                                & \textbf{Likert used for factual correctness / completness} \\\midrule
\textit{Nature \& Subjournals}                                                                                                 &                                                                                                                                                                                                                                                                                                                                                                                                                                                                                                                                                                                                                                                                                                                     &                                                                                                                                                                                           &                                                   \\
Harnessing ChatGPT and GPT-4 for evaluating the rheumatology questions of the Spanish access exam to specialized medical training & The medical experts evaluated the clinical reasoning of the chatbots followed in each of the responses. Their evaluation was based on a 1–5 scale, where a score of 5 indicates that the reasoning was entirely correct and flawless, while a score of 1 signifies that the reasoning was inconsistent or contained significant errors.                                                                                                                                                                                                                                                                                                                                                                             & 1) overall correctness.                                                                                                                                                                   & Yes                                               \\
A pilot study on the efficacy of GPT-4 in providing orthopedic treatment recommendations from MRI reports                         & {[}in Table 2{]} Likert scales used for (a) Treatment recommendations are clinically useful and relevantTreatment recommendations are clinically useful and relevant (b) Treatment recommendations are based on scientific and clinical evidence (c) The overall quality of the treatment recommendations                                                                                                                                                                                                                                                                                                                                                                                                           & 1) overall quality; 2) based on evidence; 3) useful and relevant; 4) up-to-date; 5) consistent.                                                                                           & Yes                                               \\
Testing the limits of natural language models for predicting human language judgements                                            & No details mentioned in the paper                                                                                                                                                                                                                                                                                                                                                                                                                                                                                                                                                                                                                                                                                   & 1) overall quality                                                                                                                                                                        & Not sure                                          \\
Availability of ChatGPT to provide medical information for patients with kidney cancer                                            & The SERVQUAL model is a research tool that assesses how five dimensions—tangibility, reliability, responsiveness, assurance, and empathy—influence customer perception. The answers to the questions are presented in a five-point Likert scale. SERVQUAL has mainly been used to evaluate the quality of medical services in hospitals and healthcare institutions.                                                                                                                                                                                                                                                                                                                                                & 1) tangibility; 2) reliability; 3) responsiveness; 4) assurance; 5) empathy.                                                                                                              & Yes                                               \\
Explaining machine learning models with interactive natural language conversations using TalkToModel                              & We evaluated the following statements along the 1–7 Likert scale at the end of the survey: Easiness: I found the conversational interface easier to use than the dashboard interface; Confidence: I was more confident in my answers using the conversational interface than the dashboard interface; Speed: I felt that I was able to more rapidly arrive at an answer using the conversational interface than the dashboard interface; Likeliness to use: based on my experience so far with both interfaces, I would be more likely to use the conversational interface than the dashboard interface in the future.                                                                                              & 1) Easiness; 2) Confidence; 3) Speed; 4) Likeliness to use.                                                                                                                               & No                                                \\
Evaluating large language models on medical evidence summarization                                                                & We systematically evaluate the quality of generated summaries via human evaluation. We propose to evaluate summary quality along several dimensions: (1) Factual consistency; (2) Medical harmfulness; (3) Comprehensiveness; and (4) Coherence. These dimensions have been previously identified and serve as essential factors in evaluating the overall quality of generated summaries. ... The order in which the summaries are presented is randomized to minimize potential order effects during the evaluation process. We utilize a 5-point Likert scale for the evaluation of each dimension.                                                                                                              & 1) Factual consistency; 2) Medical harmfulness; 3) Comprehensiveness; 4) Coherence.                                                                                                       & Yes                                               \\
A large-scale comparison of human-written versus ChatGPT-generated essays                                                         & The questionnaire covers the seven categories relevant for essay assessment shown below: Topic and completeness; Logic and composition; Expressiveness and comprehensiveness; Language mastery; Complexity; Vocabulary and text linking; Language constructs. ... These categories are used as guidelines for essay assessment 44 established by the Ministry for Education of Lower Saxony, Germany. For each criterion, a seven-point Likert scale with scores from zero to six is defined, where zero is the worst score (e.g. no relation to the topic) and six is the best score (e.g. addressed the topic to a special degree). The questionnaire included a written description as guidance for the scoring. & 1) Topic and completeness; 2) Logic and composition; 3) Expressiveness and comprehensiveness; 4) Language mastery; 5) Complexity; 6) Vocabulary and text linking; 7) Language constructs. & Yes                                               \\
People devalue generative AI’s competence but not its advice in addressing societal and personal challenges                       & Participants were asked to rate the author competence on three items: The author is knowledgeable of the subject; The text is credible; I intend to follow the provided recommendations.                                                                                                                                                                                                                                                                                                                                                                                                                                                                                                                            & 1) knowledgeable 2) credible 3) willing to follow.                                                                                                                                        & Not sure                                          \\
Quality of information and appropriateness of ChatGPT outputs for urology patients                                                & The responses generated by ChatGPT were then compared to those provided by a board-certified urologist who was blinded to ChatGPT’s responses and graded on a 5-point Likert scale based on accuracy, comprehensiveness, and clarity as criterias for appropriateness.                                                                                                                                                                                                                                                                                                                                                                                                                                              & 1) accuracy; 2) comprehensiveness; 3) clarity.                                                                                                                                            & Yes                                              \\\bottomrule
\end{tabular}
\caption{Papers containing ``GPT'', ``medical'' and ``Likert'' from Nature and Subjournals. }
\end{table*}

\begin{table*}[htb]
\centering\scriptsize
\renewcommand{\arraystretch}{2}
\begin{tabular}{p{3cm}p{7cm}p{2.5cm}p{1.5cm}}
\toprule
\textbf{Title}                                                                                                                             & \textbf{Relevant quote from paper}                                                                                                                                                                                                                                                                                                                                                                                                                                                                                                                                                                                                                                                                                           & \textbf{Dimensions}                                                                                                                                                                                & \textbf{Likert used for factual correctness / completness} \\\midrule
\textit{Lancet}                                                                                                                                                    &                                                                                                                                                                                                                                                                                                                                                                                                                                                                                                                                                                                                                &                                                                                                                                                                    &          \\
Assessing the potential of GPT-4 to perpetuate racial and gender biases in health care: a model evaluation study                                                      & Factual correctness and humanness of letters were assessed by two independent clinicians using a Likert scale ranging from 0 to 10, with 0 representing completely incorrect or inhuman and 10 representing completely correct and human.                                                                                                                                                                                                                                                                                                                                                                      & two dimensions: correctness and humanness.                                                                                                                         & Yes      \\
\textit{Journal of Medical Internet Research (JMIR)}                                                                                                                                                      &                                                                                                                                                                                                                                                                                                                                                                                                                                                                                                                                                                                                                &                                                                                                                                                                    &          \\
Putting ChatGPT’s Medical Advice to the (Turing) Test: Survey Study                                                                                                   & Participants were also asked about their trust in chatbots’ functions in patient-provider communication, using a Likert scale from 1-5. ... On average, responses toward patients’ trust in chatbots’ functions were weakly positive (mean Likert score 3.4 out of 5), with lower trust as the health-related complexity of the task in the questions increased.                                                                                                                                                                                                                                               & 1) trustworthy                                                                                                                                                     & Not sure \\
A Generative Pretrained Transformer (GPT)–Powered Chatbot as a Simulated Patient to Practice History Taking: Prospective, Mixed Methods Study                         & To assess how our participants perceived the simulated patient, we used the Chatbot Usability Questionnaire (CUQ). This 16‑item questionnaire measures the personality, user experience, error management, and onboarding of a chatbot and has recently been validated. ... For the CUQ, we provided relative numbers of Likert categories.                                                                                                                                                                                                                                                                    & 1) Personality; 2) User Experience; 3) Error Handling; 4) Onboarding; 5) Other                                                                                     & Yes      \\
Health Care Trainees’ and Professionals’ Perceptions of ChatGPT in Improving Medical Knowledge Training: Rapid Survey Study                                           & The questionnaire was designed according to the Kirkpatrick model, with four dimensions to understand the thoughts of the students: (1) perceived knowledge acquisition (KA), (2) perceived training motivation (TM), (3) perceived training effectiveness (TE), and (4) perceived training satisfaction (TS). Three experts reviewed and edited the questionnaire, which has 12 questions, including 1 open-ended question. A 5-point Likert scale was adopted for all questionnaire items (from 1=strongly disagree to 5=strongly agree).                                                                    & 1) perceived knowledge acquisition (KA); 2) perceived training motivation (TM); 3) perceived training effectiveness (TE); 4) perceived training satisfaction (TS). & No       \\
Assessing Health Students' Attitudes and Usage of ChatGPT in Jordan: Validation Study                                                                                 & The survey tool was created based on the TAM framework. It comprised 13 items for participants who heard of ChatGPT but did not use it and 23 items for participants who used ChatGPT. ... Each item was evaluated on a 5-point Likert scale with the following responses: strongly agree scored as 5, agree scored as 4, neutral/no opinion scored as 3, disagree scored as 2, and strongly disagree scored as 1. The scoring was reversed for the items implying a negative attitude toward ChatGPT.                                                                                                         & 13 items                                                                                                                                                           & No       \\
Increasing Realism and Variety of Virtual Patient Dialogues for Prenatal Counseling Education Through a Novel Application of ChatGPT: Exploratory Observational Study & Sentences were then appraised by a neonatologist for realism, relevance, and usability for virtual prenatal counseling simulations. Each metric used a 5-point Likert scale, ranging from 1 (the lowest) to 5 (the highest).                                                                                                                                                                                                                                                                                                                                                                                   & 1) realism; 2) relevance; 3) usability.                                                                                                                            & Not sure \\
ChatGPT Versus Consultants: Blinded Evaluation on Answering Otorhinolaryngology Case–Based Questions                                                                  & The questions were answered by both ORL consultants and ChatGPT 3. ORL consultants rated all responses, except their own, on medical adequacy, conciseness, coherence, and comprehensibility using a 6-point Likert scale.                                                                                                                                                                                                                                                                                                                                                                                     & 1) medical adequacy; 2) conciseness; 3) coherence; 4) comprehensibility                                                                                            & Not sure \\
Evaluation of GPT-4’s Chest X-Ray Impression Generation: A Reader Study on Performance and Perception                                                                 & In a blind randomized reading, 4 radiologists rated the impressions based on ``coherence'', ``factual consistency'', ``comprehensiveness'', and ``medical harmfulness'', which were used to generate a radiological score based on a 5-point Likert scale of each dimension.                                                                                                                                                                                                                                                                                                                                   & 1) coherence; 2)factual consistency; 3) comprehensiveness; 4) medical harmfulness.                                                                                 & Yes      \\
Investigating the Impact of User Trust on the Adoption and Use of ChatGPT: Survey Analysis                                                                            & We developed 2 latent constructs based on the question (predictors): Trust and Intent to Use. Participant responses to all the questions were captured using a 4-point Likert scale ranging from 1=strongly disagree to 4=strongly agree. The Actual Use factor, the outcome variable, was captured using a single-item question capturing the frequency of use ranging from 1=once a month to 4=almost every day.                                                                                                                                                                                             & 1) trust; 2) intent to use.                                                                                                                                        & Not sure \\
Exploring the Possible Use of AI Chatbots in Public Health Education: Feasibility Study                                                                               & Medical students’ feedback was collected anonymously at the end of the training experience through a 3-item questionnaire with a Likert scale (1 to 10) regarding their general satisfaction, willingness to repeat the experience, and ease of use of the tool. In particular, the scale of the 3 items can be translated as follows:; Item 1: 1=``dissatisfied with the experience'', 10=``very satisfied''; Item 2: 1=``I would not repeat the experience'', 10=``I would definitely repeat the experience''; Item 3: 1=``the tool is too difficult to be used'', 10=``the tool was very easy to be used''. & 1) satisfy; 2) intent to use; 3) easy to use.                                                                                                                      & No      \\\bottomrule
\end{tabular}
\caption{Papers containing ``GPT'' and ``Likert'' from Lancet and JMIR.}\label{app:tab:GPTlikert}
\end{table*}

\end{document}